\documentclass[lettersize,journal]{IEEEtran}
\usepackage{cite}
\def\BibTeX{{\rm B\kern-.05em{\sc i\kern-.025em b}\kern-.08em
    T\kern-.1667em\lower.7ex\hbox{E}\kern-.125emX}}
\ifCLASSINFOpdf

\else

\fi

\usepackage{times}  
\usepackage{helvet} 
\usepackage{courier}  
\usepackage[hyphens]{url}  
\usepackage{graphicx} 
\usepackage{graphicx}  
\usepackage[pagebackref=true,breaklinks=true,colorlinks,bookmarks=false]{hyperref}
\hypersetup{
    colorlinks=true,
    filecolor=magenta,      
    urlcolor=cyan,
}

\usepackage{subcaption}
\usepackage{amsmath,graphicx}

\usepackage{amsmath}
\usepackage{xcolor,soul,framed} 
\usepackage{xspace}
\usepackage{array}
\usepackage{eqparbox}
\usepackage{url}
\usepackage{longtable}
\usepackage{lipsum}
\usepackage{blindtext}
\usepackage{makecell}
\usepackage{mathtools}
\usepackage{commath}
\usepackage{multirow}
\usepackage{tabularx}
\usepackage[normalem]{ulem}
\usepackage{xspace}
\usepackage{algorithm}
\usepackage{algpseudocode}
\usepackage{amsfonts}

\newcolumntype{C}{>{\hsize=\dimexpr0.5\hsize+8\tabcolsep+\arrayrulewidth\centering\relax}X}

\DeclareMathOperator{\LSTM}{LSTM}

\DeclareMathOperator{\Tr}{Tr}
\DeclareMathOperator{\Log}{Log}
\DeclareMathOperator{\Exp}{Exp}
\DeclareMathOperator{\rank}{rank}
\DeclareMathOperator{\Vect}{Vect}
\DeclareMathOperator{\Upper}{Upper}
\DeclareMathOperator{\cref}{ref}
\DeclarePairedDelimiter\floor{\lfloor}{\rfloor}

\begin{document}

\title{Spatio-Temporal EEG Representation Learning on Riemannian Manifold and Euclidean Space}

\author{Guangyi~Zhang~and~Ali~Etemad,~\IEEEmembership{Senior~Member,~IEEE}
\thanks{G. Zhang and A. Etemad are with the Department of Electrical and Computer Engineering $\&$ Ingenuity Labs Research Institute, Queen's University, Kingston, Canada, K7L 3N9 e-mail: \{guangyi.zhang, ali.etemad\}@queensu.ca}
}

\maketitle

\begin{abstract}
We present a novel deep neural architecture for learning electroencephalogram (EEG). To learn the spatial information, our model first obtains the Riemannian mean and distance from spatial covariance matrices (SCMs) on a Riemannian manifold. We then project the spatial information onto a Euclidean space via tangent space learning. Following, two fully connected layers are used to learn the spatial information embeddings. Moreover, our proposed method learns the temporal information via differential entropy and logarithm power spectrum density features extracted from EEG signals in a Euclidean space using a deep long short-term memory network with a soft attention mechanism. To combine the spatial and temporal information, we use an effective fusion strategy, which learns attention weights applied to embedding-specific features for decision making. We evaluate our proposed framework on four public datasets across three popular EEG-related tasks, notably emotion recognition, vigilance estimation, and motor imagery classification, containing various types of tasks such as binary classification, multi-class classification, and regression. \textcolor{black}{Our proposed architecture outperforms other methods on SEED-VIG, and approaches the state-of-the-art on the other three datasets (SEED, BCI-IV 2A, and BCI-IV 2B), showing the robustness of our framework in EEG representation learning.} The source code of our paper is publicly available at \href{https://github.com/guangyizhangbci/EEG\_Riemannian}{https://github.com/guangyizhangbci/EEG\_Riemannian}.

\end{abstract}

\IEEEpeerreviewmaketitle

\section{Introduction} \label{sec:intro}
Brain computer interfaces (BCI) enable communication between users and computers through learning and interpreting brain activity, for example, brain signals and neuroimaging \cite{zheng2015investigating}. Non-invasive technologies such as electroencephalogram (EEG), functional magnetic resonance imaging \cite{phan2002functional}, functional near-infrared spectroscopy \cite{plichta2011auditory}, and magnetoencephalography \cite{lal2005brain}, have been widely used for BCI. Among the above-mentioned technologies, EEG is one of the most popular due to portability and low cost, high temporal resolution, and the ability to provide real-time monitoring. 

EEG-based BCI systems have been widely used in many application areas. For example, BCI can enable users to move virtual/digital objects on screens via imagining specific movements (e.g., left hand and right hand) \cite{leeb2008bci}. BCI can also help identify users' affective states (e.g., happy, sad, or neutral) through learning neural patterns when consuming emotionally charged content \cite{zheng2015investigating}. Furthermore, BCI can provide early detection of fatigue or other impairments through real-time monitoring of the brain activity \cite{zheng2017multimodal}.

Various approaches have been proposed and implemented for brain-computer interaction through learning discriminative task-relevant features from EEG signals. For example, spatial filtering has been one of the \textcolor{black}{most common techniques} used to explore the features that \textcolor{black}{contain the optimal amount} of variance with respect to different tasks \cite{ramoser2000optimal}. Statistical models have also been used to investigate linear relationships between EEG features and output labels \cite{thompson2005canonical,zheng2016multichannel}. 
Numerous machine learning techniques have been implemented to model the nonlinear relationships encountered in EEG-based classification tasks \cite{lotte2007review}. Lastly, deep learning techniques have considerably improved the performance of EEG-based BCI systems in recent years \cite{lotte2018review}.

Given the complexity and high dimensionality of EEG, most existing solutions are often unable to learn the non-linearities observed in high-dimensional multi-channel EEG manifolds and extracted representations. As a result, most existing EEG-related works propose pipelines customized for particular classification or regression tasks. Thus, developing frameworks for EEG representation learning that generalize well across different BCI tasks (e.g., motor imagery (MI) classification, emotion recognition, and vigilance estimation) remains challenging \cite{ko2021multi,lawhern2018eegnet}.

Thus far, many BCI solutions rely on spatial covariance matrices (SCMs) computed from raw multi-channel EEG to learn spatial information. Euclidean metric learning such as the distance between two SCMs and averaging of SCMs have been widely used for EEG signal classification \cite{ang2012filter,yger2015averaging,kalunga2015euclidean}. Euclidean metric learning, on the other hand, suffers from poor signal classification performance due to two problems. First, EEG often suffers from the linear mixing effect due to volume conduction \cite{van1998volume,khadem2014quantification,palva2018ghost,imperatori2019eeg}, resulting in the poor estimation of the distance between two SCMs. Second, SCMs of raw EEG are symmetric positive definite (SPD). 
The calculation of Euclidean mean of SPD matrices may be inaccurate since the determinant of the mean value of SPD matrices can be strictly larger than the determinant of any of the SPD matrices (also known as the swelling effect) \cite{arsigny2007geometric}. This in turn will often result in poor classification performance \cite{kalunga2015euclidean}.

To overcome the above-mentioned issues in BCI applications, a Riemannian metric learning method was recently applied on SCMs and enhanced the EEG classification performance \cite{sabbagh2019manifold,lotte2018review}. Riemannian distance between any two full rank SCMs, unlike the Euclidean distance, is \textbf{affine-invariant} \cite{sabbagh2019manifold}. If any linear transformation is applied on EEG signals, this affine-invariance property will allow the distance between the two SCMs of EEG signals to remain unchanged. As a result, the linear mixing effect of EEG will be minimized when Riemannian distance is used for EEG classification \cite{pennec2006riemannian}. Furthermore, according to \cite{kalunga2015euclidean}, no swelling effect exists during the estimation of Riemannian mean.

In this paper, we aim to to provide a generalized solution for BCI applications by effectively learning EEG representations. 
Our method first learns spatial information on a Riemannian manifold and temporal information in a Euclidean space individually, then fuses the learned representations effectively. 
Our spatial information includes the distance between two SCMs as well as the mean of SCMs, while our temporal information contains EEG features extracted and learned from signals in consecutive time periods. We overcame the following challenges during this process:

\noindent \textbf{(1) Spatial information learning on a Riemannian manifold.}
Because SPD matrices belong to a Riemannian manifold rather than a Euclidean space \cite{arsigny2007geometric,kalunga2015euclidean}, it is important to project spatial information learned from a Riemannian manifold \cite{arsigny2007geometric} onto a Euclidean space. For instance, direct spatial feature learning on a Riemannian Manifold is not appropriate because important information 
may be lost during the learning process with Euclidean metrics. To address this challenge, we project the spatial information from a Riemannian manifold onto a Euclidean space via tangent space learning before feeding the learnt embeddings to a deep learning architecture, ensuring that the \textcolor{black}{important information is} preserved
\cite{arsigny2007geometric}. More importantly, the information transformed to Euclidean space may end up having a high dimensionality, which in turn can result in overfitting. We overcome this issue by resorting to dropout layers within our model.

\noindent \textbf{(2) Riemannian metric learning on SCMs.} 
The SCMs of raw multi-channel EEG are often SPD since the signals in each channel cannot be strictly expressed as the linear combination of the others (also known as full rank) \cite{sabbagh2019manifold}. However, the artefact suppression steps performed during EEG pre-processing 
may discard portions of information from multi-channel EEG \cite{uusitalo1997signal}. Insufficient data may also result in poor estimation of SCMs \cite{engemann2015automated}. This, in turn, may result in rank deficiency of EEG, leading the SCMs to become symmetric positive semi-definite (SPSD) which is no longer suitable for applying Riemannian metrics \cite{sabbagh2019manifold}. To tackle this problem, we employ principle component analysis (PCA) to project the SCMs from SPSD to SPD via dimensionality reduction, enabling the Riemannian metric learning on SCMs \cite{sabbagh2019manifold}, while capturing most of the variance.

\noindent \textbf{(3) Feature fusion of the Riemannian spatial and Euclidean temporal information}. Combining two different representations of the input, spatial information on a Riemannian manifold and temporal information in a Euclidean space, is challenging given their different geometric structures \cite{kalunga2015euclidean}. 
Moreover, each representation contributes differently to the final task at hand, requiring a learned fusion strategy for optimal performance. To solve these challenges, we apply an encoder for each information stream and effectively learn both the original and encoded information with a soft attention mechanism.

We build the solutions above in a single end-to-end deep architecture. 
To show the robustness of our proposed network, we evaluate the proposed architecture on four public datasets based on the following considerations: \textit{i}): covering different application areas of EEG-based BCI such as emotion classification, vigilance estimation, and MI classification; \textit{ii}): including problems with both binary and multi-class classification (2, 3, or 4 classes); \textit{iii}): the tasks containing both continuous label prediction (regression) as well as classification; and \textit{iv}): the tasks consisting of different distributions and number of EEG channels.   

In summary, we make the following contributions:
\begin{itemize}
\item We propose a \textit{novel framework} for EEG representation learning based on learning spatial information on a Riemannian manifold and temporal information in a Euclidean space followed by an effective fusion strategy. 
\item We test the proposed framework on \textit{three different EEG-related problem domains} namely emotion recognition, motor-imagery classification, and vigilance estimation, using \textit{four widely used public datasets}.
\item \textcolor{black}{Our method performs well in all the experiments, approaching the state-of-the-art in three datasets and outperforming the best results of existing works in one dataset, setting new state-of-the-art.}
\end{itemize}

The rest of this paper is organized as follows. In Section \ref{sec: related work}, we provide an overview of related work on EEG-based BCI applications in the three different application areas namely emotion recognition, MI classification, and vigilance estimation. Section \ref{sec: proposed architecture} gives a systematic description of the proposed architecture including feature extraction and learning for both spatial correlations and temporal dependencies, as well as the fusion strategy used. In Section \ref{sec: experiments}, we give a description of all the datasets, implementation details, and evaluation protocols. We further discuss the results and perform ablation studies. Section \ref{sec: summary} presents the summary and conclusions of this paper.

\section{Related Work}\label{sec: related work}
In this section, we summarize the related work on the problem domains studies in this paper. First, we group and study on emotion recognition and vigilance estimation papers together as many common techniques have been used for these two areas. Next, we provide an overview of the related work on motor-imagery classification.

\subsection{Emotion Recognition and Vigilance Estimation}
Recently, numerous EEG-based solutions have been proposed for emotion recognition. Pipelines usually consist of feature extraction followed by a classification or regression network \cite{lotte2007review}. In these approaches, a critical step is often the selection of powerful features from noisy raw EEG signals due to the non-linear and non-stationary nature of EEG \cite{zhang2019classification}. As an example, \textcolor{black}{differential entropy} (DE) has been recently reported as an effective and robust feature for emotion classification and vigilance regression models \cite{zheng2015investigating,zheng2017multimodal}. Successive to feature extraction, various types of algorithms have been successfully exploited for the classification/regression tasks. For instance, group sparse canonical correlation analysis (GSCCA) was proposed to model the \textit{linear} relationship between extracted features (including DE) and output labels \cite{zheng2016multichannel}. To investigate the non-linearities in the aforementioned extracted features, several classical machine learning methods such as k-nearest neighbor (kNN) \cite{zheng2015investigating}, linear regression (LR) \cite{zheng2015investigating}, graph regularized sparse linear regression (GRSLR) \cite{li2019eeg}, support vector machine (SVM) \cite{zheng2015investigating}, and random forest (RF) \cite{li2019novel} have been used for EEG-based emotion classification. Support vector regression (SVR) was employed in \cite{zheng2017multimodal} to predict continuous values for a regression formulation of the problem. To better learn the extracted features, \textcolor{black}{feed-forward artificial neural networks (ANNs) such as graph regularized extreme learning machine (GELM) \cite{zheng2017identifying} have been used to improve the performance.} 

In order to explore the most discriminative and task-relevant features, deep learning frameworks were applied. For instance, deep belief network (DBN) was employed to extract high-level representations through deep hidden layers \cite{zheng2015investigating}. Double-layered neural network with subnetwork nodes (DNNSN) was adopted for predicting vigilance labels \cite{wu2018regression}. Convolutional neural network (CNN) along with capsule attention \cite{zhang2021capsule} was adopted to learn spatiotemporal EEG information. Spatial-temporal recurrent neural network (STRNN) \cite{zhang2018spatial} and long short-term memory (LSTM) network \cite{zhang2016continuous} have been employed to learn the temporal information embedded in the EEG time-series. \textcolor{black}{A three-dimensional convolutional attention neural network was introduced to capture the dynamic interactions and internal spatial relationships among EEG signals over continuous time periods \cite{liu20213dcann}. Similarly, an end-to-end spatio-temporal demographic network was proposed to integrate both spatial and temporal EEG information using single-link hierarchical clustering \cite{li2022sstd}.} Domain adaptation network (DAN) \cite{li2018cross} was recently exploited to achieve better performance with utilizing prior knowledge of data distribution in the target domain. 
\textcolor{black}{DAN aims to reduce the effects of domain shift by leveraging the information from the source domain (train set) and adapting it to the target domain (test set). This could be achieved by aligning the distributions of the source and target domains in a shared latent space.}
Bi-hemispheres domain adversarial neural network (BiDANN) was proposed to minimize the domain shift between training and testing data through a discriminator \cite{li2018novel}. Bi-hemispheric discrepancy model (BiHDM) was proposed to improve the performance based on the architecture of an RNN and DAN through learning domain-invariant features from two brain hemispheres \cite{li2019novel}. A similar approach, regional to global brain-spatial-temporal neural network (R2G-STNN), explored spatial-temporal features through bidirectional LSTM (BiLSTM) and decreased the domain-shift through training a discriminator \cite{li2019regional}. Recently, LSTM has also been used to explore variational pathway reasoning (VPR) \cite{zhang2020variational}, and has approached state-of-the-art performance by firstly employing the RNN network to explore the between-electrode dependencies, thus encoding pathways generated from random walk. Then, it chose salient pathways with the most important pair-wise connections via scaling factors as well as pseudo-pathways. Graph neural networks (GNN) such as dynamical graph convolutional neural networks (DGCNN) \cite{song2018eeg}, and regularized graph neural networks (RGNN) \cite{zhong2020eeg} have recently been utilized to explore the topological structure of EEG electrodes as well as inter-channel relationships by learning graph connections, approaching state-of-the-art results.

\subsection{EEG Motor Imagery Classification}
Many works on EEG-based MI research rely on CSP as the spatial filtering method for classification \cite{ang2012filter}. Filter bank common spatial pattern (FBCSP) which decomposes EEG to few sub-frequency bands before the use of CSP has also been frequently used for feature extraction \cite{ang2012filter}. For example, in binary classification problems (e.g., left hand Vs. right hand), CSP filters maximize the variance of EEG trials from the left-hand class while minimizing the variance of the EEG trial from the right-hand class, through applying simultaneous diagonalization on two covariance matrices from both classes whose eigenvalues are summed to one \cite{ang2012filter}. CSP filters have also been used in multi-class MI tasks, mainly through one-versus-rest and one-versus-one strategies \cite{tangermann2012review,ghaheri2013extracting,hersche2018fast}. Several classifiers such as SVM, na\"ive Bayes (NB), RF, and linear discriminant analysis (LDA) have been reported with considerable results while using CSP or FBCSP as the feature extraction method \cite{tangermann2012review,ghaheri2013extracting,hersche2018fast,bentlemsan2014random}. 
For multi-class classification, extended sequential adaptive fuzzy inference system (ESAFIS) proposed in \cite{rong2018incremental} shows better results on learning CSP features in comparison to other classifiers such as SVM. To better explore the energy features through CSP, deep learning techniques such as CNN with average pooling and the parallel combination of multi-layer perceptron (MLP) and CNN has been used and achieved better results compared to SVM \cite{sakhavi2015parallel}. 

Lastly, end-to-end deep learning approaches have recently been adopted in MI classification. CNN-based architectures such as EEGNet have been directly implemented on raw EEG data \cite{hersche2020compressing}. In another direct approach, capsule networks (CapsNet) have been applied on spectrogram of EEG data, achieving considerable results \cite{ha2019decoding}. \textcolor{black}{Similar to emotion recognition, the integration of spatial and temporal EEG information plays a crucial role in MI pattern recognition. For instance, a CNN-based end-to-end framework utilized two-dimensional convolution blocks to extract both spatial and temporal features from EEG signals \cite{tang2023motor}. Furthermore, a framework based on graph convolution was designed to leverage not only spatial and temporal features, but also their interactions for MI-EEG decoding \cite{ma2022spatio}.}

\section{Proposed Architecture}\label{sec: proposed architecture}
In the following context, the notations used are described as  follow: `a' represents a scalar, `$\mathbf{a}$' represents a vector, `$\mathbf{A}$' represents a matrix, `$\mathcal{A}$' represents a differentiable manifold.    

\subsection{Solution Overview}
We design a novel architecture for learning spatio-temporal EEG representations. Initially, we apply a filter bank on EEG. In order to learn the spatial information, we first compute SCMs of multiple frequency sub-bands. Then, to tackle the possible rank deficiency caused by artefact suppression, we employ PCA to ensure SCMs are in the space of SPD, thus enabling the use of affine-invariant Riemannian distance. Next, we apply the Riemannian distance on the SCMs and estimate the Riemannian mean. Following, we map the spatial information of SCMs on Riemannian manifold to the feature vectors in Euclidean space via tangent space learning, with the Riemannian mean as the reference. Lastly, we use two fully connected (FC) layers to learn the spatial information embedded in the feature vectors.

In order to obtain temporal information in Euclidean space, we employ a three-layer LSTM network with attention to learn the temporal dependencies of entropy and frequency features extracted from the same EEG frequency sub-bands as those used in the Riemannian pipeline. Next, we feed forward the temporal information learned from the attention mechanism to an FC layer to obtain latent representations.

Following that, to learn the mutual and selective information embedded in the latent spatial and temporal representations in Euclidean space, we exploit a fusion strategy to obtain a final embedding suitable for various classification or regression tasks. 

\subsection{Data Pre-processing}
To keep consistent with the related work using the same datasets, EEG sampling rates were downsampled from $1000 Hz$ to $200 Hz$ for emotion and vigilance datasets while being kept unchanged at $250Hz$ for both MI datasets. For each of the four datasets, the EEG signals were band-pass filtered between $0.5-70$ \textit{Hz} using a $5^{th}$ order Butterworth filter to lower artifacts. Then, a notch filter at $50$ \textit{Hz} was applied to reduce power line noise. \textcolor{black}{Signal amplitudes were re-scaled} to the range of $[-1, 1]$ through min-max normalization so that the data discrepancy across different recording sessions was decreased for each subject.

\subsection{Temporal Feature Processing}
\subsubsection{Feature Extraction}
Two types of features namely logarithm power spectrum density (PSD) and DE are extracted. PSD is defined in Eq. \ref{equation: psd} and DE of EEG time-series $X$ with a Gaussian distribution is shown in Eq. \ref{equation: de} respectively. To avoid spectral leakage, frequency domain features are extracted through short-time Fourier transform (STFT) using the periodogram method. $1$-second Hanning windows are used with an overlap of $50\%$ offering $L$ windows ($L = \floor*{2\times T - 1}$, where $T$ is the length of each EEG segment) for feature extraction.
\begin{equation}\label{equation: psd}
S_{xx}(\omega)=\lim_{T\to\infty}E\Big[|\hat{X}(\omega)|^{2}\Big].
\end{equation}
\begin{equation}\label{equation: de}
DE = \frac{1}{2} \log{(2\pi e \sigma^{2})}, \hspace{3mm}  X \sim N(\mu, \sigma^{2}).
\end{equation}

\subsubsection{LSTM Network with Attention}
LSTM is a type of RNN that enables the learning of both long and short-term dependencies from sequential data (e.g., text, audio, and bio-signals) while addressing gradient exploding and vanishing problems \cite{hochreiter1997long}. LSTM networks have been recently successfully implemented on EEG signals in BCI tasks and achieved notable results \cite{zhang2016continuous,zhang2019classification}. Similarly, in our experiments, following feature extraction, concatenated task-relevant EEG features (DE and logarithm PSD) from different frequency sub-bands (total number of $H$) in different windows are fed into $L$ corresponding LSTM cells (also called time-steps). Then, information in different LSTM cells ($\mathbf{s}_i$) are learned through deciding which part to remember or forget through weights updated during network training \cite{zhang2019classification}. As shown in Eq. \ref{h_i}, $\mathbf{h}_i$ is the generated output of the hidden states at each time-step $i$ which are passed forward to the the LSTM cell of the next LSTM layer for higher level feature representation learning. 
\begin{equation}\label{h_i}
\mathbf{h_i} = \LSTM(\mathbf{s_{i}}), i\in [1,L],
\end{equation}

To improve the capability of handling temporal information, deep LSTM architectures followed by soft attention or capsule attention mechanisms have been lately implemented showing great performance on different EEG-related classification or regression tasks \cite{zhang2019classification,zhang2021capsule}. Compared to a conventional LSTM network that only considers the last hidden state $h_L$ as the network output, a soft attention mechanism evaluates the importance of all output information ($\{\mathbf{h}_i\}_{i=1}^L$) from the last LSTM layer by assigning trainable attention weights $\alpha_i$ applied on each $h_i$ as shown in Eq. \ref{u_i} and \ref{alpha_i}. Thus, more task-relevant information can be obtained by focusing on certain time-steps through optimizing attention weights. The equations are presented as follows: 
\begin{equation}\label{u_i}
\mathbf{u}_i = tanh(\mathbf{W}\mathbf{h}_i + \mathbf{b}), 
\end{equation}
\begin{equation}\label{alpha_i}
\mathbf{\alpha_i} = \frac{exp(\mathbf{u}_i)}{\sum_j{exp(\mathbf{u}_j)}},
\end{equation}
\begin{equation}\label{v_i}
\mathbf{v} = \sum_i{\alpha_i \mathbf{h}_i},
\end{equation}
where vector $\mathbf{v}$ is the output of the LSTM network with attention, \textcolor{black}{and $\mathbf{W}$ and $\mathbf{b}$ are the trainable parameters}. Accordingly, in our architecture, we employ a soft attention mechanism following a three-layer LSTM network to help focus on the most discrepant higher level features in different time-steps.

\begin{figure}[!ht]
    \begin{center}
    \includegraphics[width=0.70\linewidth]{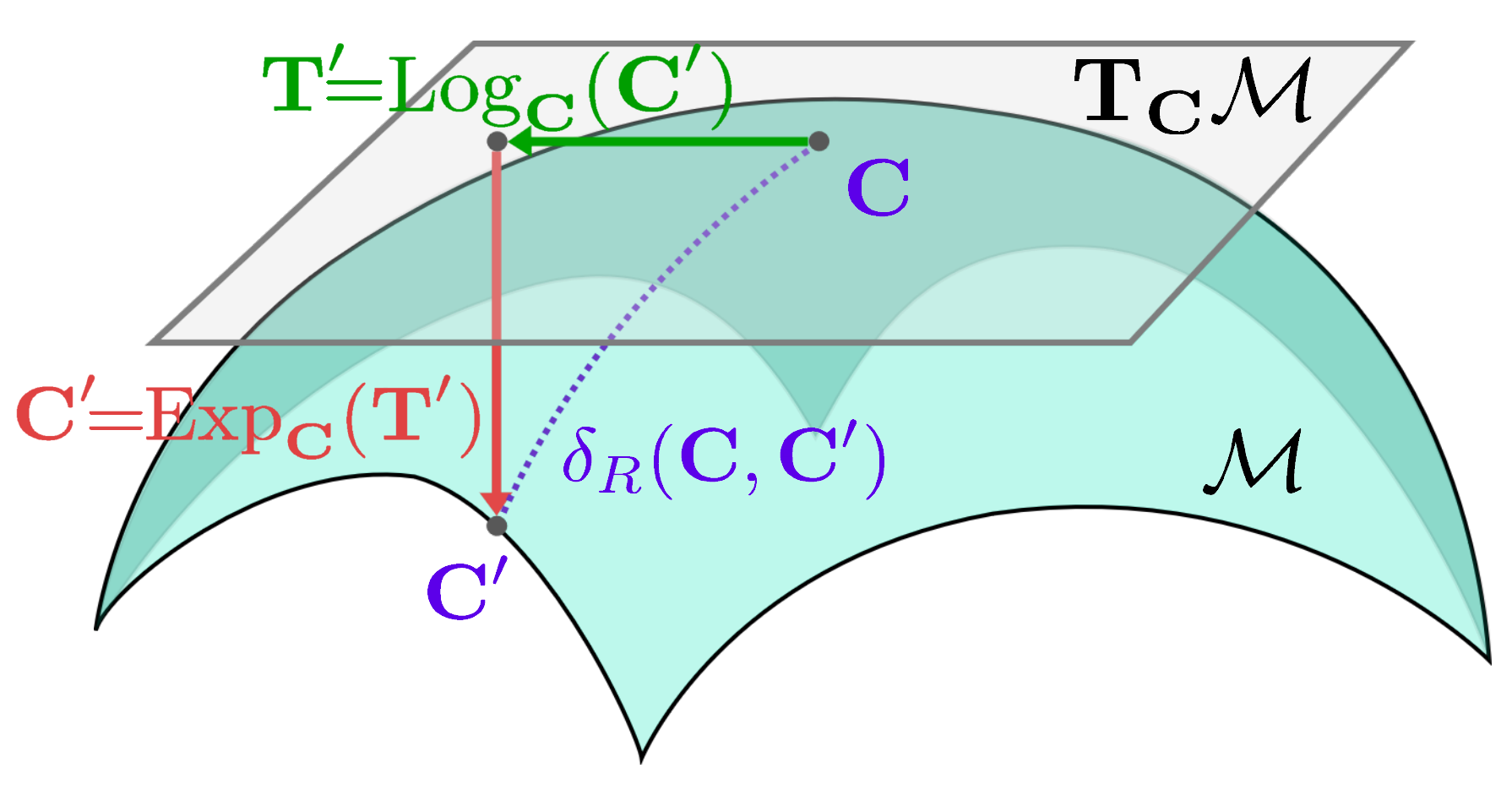} 
    \end{center}
\caption{The Concept of Riemannian Manifold. \textcolor{black}{It demonstrates estimation on distance between two SPD matrices (C and C') on Riemannian manifold using the tangent space learning technique.}}
\label{fig-manifold}
\end{figure}

\subsection{Spatial Feature Processing}
\textbf{Background:} 
As mentioned earlier in the Introduction, SCMs of raw EEG are SPD matrices on a Riemannian manifold \cite{sabbagh2019manifold}. Riemannian geometry is employed to better learn and manipulate the SPD matrices, in order to capture spatial information. Recent studies show that Riemannian approach achieves better performance than CSP approaches using the same classifier in BCI applications \cite{barachant2013classification}. Other Riemannian approaches using minimum distance to Riemannian mean (MDRM) and tangent space LDA (TSLDA) as classifiers consistently outperformed CSP methods in different EEG classification tasks \cite{xie2016motor}. Local isometric embedding (LIE) was proposed based on using tangent space to better learn the features through dimensionality reduction, compared with MDRM classifier and tangent space followed by an SVM classifier \cite{li2019motor}. A very recent study claimed that artefact-suppression reduced the robustness of Riemannian-based approaches \cite{sabbagh2019manifold}. Next, we briefly introduce Riemannian geometry.

\textbf{Riemannian Geometry:}
Let $\mathcal{M}$ be a differentiable manifold with $G$ dimensions. As shown in Figure \ref{fig-manifold}, $\mathbf{T}_\mathbf{C}\mathcal{M}$ denotes the tangent space (also called derivative) of $\mathcal{M}$ at $\mathbf{C} \in \mathcal{M}$. 

The inner product of two tangent vectors ($\mathbf{T}_1,\mathbf{T}_2 \in \mathbf{T}_\mathbf{C}\mathcal{M}$) is defined as \cite{tuzel2008pedestrian}:
\begin{equation}\label{inner_product}
\big < \mathbf{T}_1, \mathbf{T}_2 \big>_\mathbf{C} = \Tr(\mathbf{T}_1\mathbf{C}^{-1}\mathbf{T}_2\mathbf{C}^{-1}),
\end{equation}
where $\Tr{(.)}$ is a trace operator. Also, the inner product introduces the norm of a tangent vector $\mathbf{T}$ as \cite{sabbagh2019manifold}:
\begin{equation}\label{inner_product_norm}
||\mathbf{T}||_\mathbf{C} = \big[\big<\mathbf{T},\mathbf{T}\big>_\mathbf{C}\big]^{1/2} = \big[\Tr(\mathbf{T}\mathbf{C}^{-1}\mathbf{T}\mathbf{C}^{-1})\big]^{1/2}.
\end{equation}

Logarithm mapping ($\Log$) in Eq. \ref{logarithm map} helps project $\mathbf{C}'$ from $\mathcal{M}$ to $\mathbf{T}'$ in $\mathbf{T}_\mathbf{C}\mathcal{M}$. Meanwhile, Exponential mapping ($\Exp$) in Eq. \ref{exponential map} is introduced to project $\mathbf{T}'$ back to $\mathbf{C}'$ as shown in following \cite{barachant2013classification}:
\begin{equation}\label{logarithm map}
\mathbf{T}'= \Log_\mathbf{C}(\mathbf{C}') = \mathbf{C}^{1/2}\log(\mathbf{C}^{-1/2}\mathbf{C}'\mathbf{C}^{-1/2})\mathbf{C}^{1/2},
\end{equation}
\begin{equation}\label{exponential map}
\mathbf{C}'= \Exp_\mathbf{C}(\mathbf{T}') = \mathbf{C}^{1/2}\exp(\mathbf{C}^{-1/2}\mathbf{T}'\mathbf{C}^{-1/2})\mathbf{C}^{1/2} ,
\end{equation}
where $\mathbf{C}, \mathbf{C}'\in \mathcal{M}, \mathbf{T}'\in \mathbf{T}_\mathbf{C}\mathcal{M}, \log{(.)}, \exp{(.)}$ are logarithm and exponential operations applied on a matrix.

Riemannian distance (also called geodesic distance) is a very important metric representing the distance of the shortest path between $\mathbf{C}$ and $\mathbf{C}'$ (shown as the curve in Figure \ref{fig-manifold}) on manifold $\mathcal{M}$. The geodesic distance ($\delta_R$) is equivalent to the length of its tangent vector \cite{forstner2003metric,tuzel2008pedestrian}, expressed as follows:
\begin{equation}\label{distance_definition}
\delta_R(\mathbf{C}, \mathbf{C}') = ||\Log_\mathbf{C}(\mathbf{C}')||_{\mathbf{C}} = ||\mathbf{T}'||_{\mathbf{C}}.
\end{equation}

In the context of this work, we denote $S_N$ = \{$\mathbf{M} \in \mathbb{R}^{N\times N}, \mathbf{M}^\top=\mathbf{M}, \mathbf{x}^\top\mathbf{M}\mathbf{x}\geq0, \forall \mathbf{x} \in \mathbb{R}^N \setminus \mathbf{0}$\} as the space of SPSD matrices. Similarly $S_N^{+}$ = \{$\mathbf{M} \in \mathbb{R}^{N\times N}, \mathbf{M}^\top=\mathbf{M}, \mathbf{x}^\top\mathbf{M}\mathbf{x}>0, \forall \mathbf{x} \in \mathbb{R}^N \setminus \mathbf{0}$\} is defined as the space of SPD matrices, $S_R = \{\mathbf{M} \in S_N, \rank(\mathbf{M})=R, R<N$ \} is the space of SPSD matrices, where $\rank(\mathbf{M})$ is the rank of a matrix, and $S_R^{+}$ = \{$\mathbf{M} \in \mathbb{R}^{R\times R}, \mathbf{M}^\top=\mathbf{M}, \mathbf{x}^\top\mathbf{M}\mathbf{x}>0, \forall \mathbf{x} \in \mathbb{R}^R \setminus \mathbf{0}$\} is the subspace of SPD matrices with full rank $R$. 

\textbf{Our Method:} To learn spatial information embedded in multi-channel EEG, we compute SCMs on filtered signals in each frequency sub-band. Suppose $\{\mathbf{X}_{i}\}_{i=1}^P \in \mathbb{R}^{N\times T}$, where $i$ denotes $i^{th}$ segment, $P$ is the EEG segment number, $N$ is the EEG channel number, and $T$ represents the number of data samples per segment. The SCMs can be estimated as $\mathbf{C}_i = \frac{1}{(T-1)}\mathbf{X}_{i}\mathbf{X}_{i}^\top$, which may be in $S_R$ after artefact suppression. PCA is then employed to project SCMs from $S_R$ to $S_R^{+}$ in order to enable the affine-invariant property during geodesic distance calculation. At last, we obtain the spatial information (geodesic distance) of SCMs in the Riemannian manifold as feature vectors in Euclidean space via tangent space learning based on the chosen reference matrix as described in the following text.

\begin{figure*}[!t]
    \begin{center}
    \includegraphics[width=1.0\textwidth]{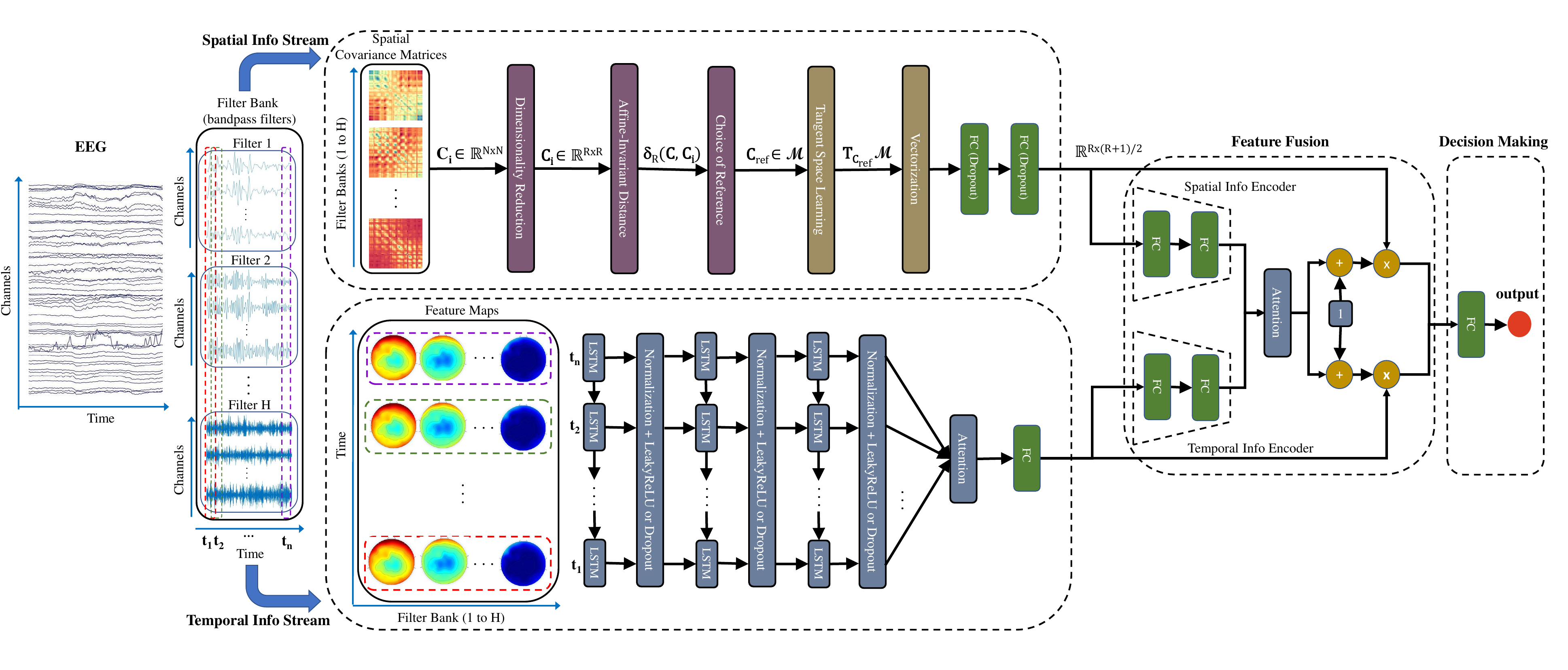} 
    \end{center}
\caption{The overview of the experiment work-flow is presented. \textcolor{black}{The raw, multi-channel EEG signals undergo preprocessing via a filter bank, followed by spatial covariance matrix learning through a Riemannian approach (indicated in the upper block). Each time step's extracted features are learned using an attention-based LSTM architecture (indicated in the lower block). Subsequently, spatial and temporal information is learned using a feature fusion method, which enables final decision-making in various tasks.}}
\label{new_architecture}
\end{figure*}

\subsubsection{Dimensionality Reduction on $\{\mathbf{C}_i\}_{i=1}^P$ in $S_R$}
Although $\mathbf{C}_i$ of raw EEG $\mathbf{X}_i$ are in $S_N^{+}$, artifact suppression may destroy some important information \cite{uusitalo1997signal} that could result in the rank deficiency of EEG data \cite{sabbagh2019manifold}, leading for $\mathbf{C}_i$ to belong to $S_R$. Wasserstein distance has been applied on $\mathbf{C}_i$ in $S_R$ \cite{sabbagh2019manifold}. However, it lacks affine-invariance in dealing with the linear mixing effect in multi-channel EEG recordings \cite{sabbagh2019manifold}. Therefore, we perform dimensionality reduction to project $\mathbf{C}_i$ to $S_R^{+}$, enabling the use of affine-invariant Riemannian distance. First, we use PCA to estimate spatial filter $\mathbf{W}$ \cite{sabbagh2019manifold}. Specifically, we sort the eigenvector matrix $\mathbf{V}$ based on descending eigenvalues, and choose $\mathbf{V}$ containing only top $R$ eigenvalues of the averaged $\{\mathbf{C}_i\}_{i=1}^P$ (Eq. \ref{Euclidean_mean}) as the spatial filter $\mathbf{W}$, thus maximizing the variance \cite{sabbagh2019manifold}. Then, we apply the spatial filter $\mathbf{W} \in \mathbb{R}^{N\times R}$ on signal $\mathbf{X}_i \in \mathbb{R}^{N\times T}$, leading to $\mathbf{X}_i'=\mathbf{W}^\top\mathbf{X}_i \in \mathbb{R}^{R\times T}$. Accordingly, the $\mathbf{C}_i$ of $\mathbf{X}_i'$ is expressed as $\mathbf{W}^\top\mathbf{C}_i\mathbf{W} \in \mathbb{R}^{R\times R}$ in $S_R^{+}$.

\subsubsection{Affine-invariant Distance of $\{\mathbf{C}_i\}_{i=1}^P$ in $S_R^{+}$}
Since $\mathbf{C}_i$ is in $S_R^{+}$ after dimensionality reduction, we apply Riemannian distance on $\mathbf{C}_i$ where the affine-invariant property of distance is illustrated as following:
\begin{equation}\label{affine-invariant}
\delta_R(\mathbf{C}, \mathbf{C}_i) = \delta_R(\mathbf{W}^\top\mathbf{C}\mathbf{W},\mathbf{W}^\top\mathbf{C}_i\mathbf{W}), 
\end{equation} 
where $\mathbf{C}, \mathbf{C}_i \in \mathbb{R}^{R\times R}, \mathbf{W}\mathbf{W}^{-1}=\mathbf{I}$, and $\mathbf{W}^\top\mathbf{X}_i'$ is the linear transform of the EEG \cite{sabbagh2019manifold}, which was proven in \cite{forstner2003metric}. The affine-invariance property (Eq. \ref{affine-invariant}) enables the geodesic distance between $\mathbf{C}$ and $\mathbf{C}_i$ in $\mathcal{M}$ to be invariant when linear transforms are applied to EEG (e.g., linear mixing effect). Unlike many approaches, we do not apply Log-Euclidean distance on $\mathbf{C}_i$ in $S_R^{+}$ due to the lack of the affine-invariance property \cite{congedo2017riemannian,kalunga2015euclidean}. Next, instead of conventional Riemannian approach that apply geodesic distances directly for classification (e.g., MDRM) \cite{barachant2011multiclass}, we preserve the geodesic information ($||\mathbf{T}_i||_\mathbf{C}$) of $\{\mathbf{C}_i\}_{i=1}^P$ as feature vectors in Euclidean space (Eq. \ref{distance_definition}) for further higher level feature learning. To achieve this, we first carefully select the reference matrix so that the spatial information on Riemannian manifold represented by geodesic distance between $\{\mathbf{C}_i\}_{i=1}^P$ and the reference matrix in $S_R^{+}$ can be projected onto the same tangent space, in order to best capture geodesic information of $\{\mathbf{C}_i\}_{i=1}^P$ in $\mathcal{M}$.

\subsubsection{Choice of Reference for $\{\mathbf{C}_i\}_{i=1}^P$ in $S_R^{+}$}
We denote $\mathbf{C}_{\cref} \in \mathcal{M}$ as the reference matrix for $\{\mathbf{C}_i\}_{i=1}^P$ during tangent space ($\mathbf{T}_{\mathbf{C}_{\cref}}\mathcal{M}$) learning. In the recent studies, the approaches using Riemannian mean ($\overline{\mathbf{C}}_R$) as the reference ($\mathbf{C}_{\cref}$) during the tangent space learning have outperformed approaches that used other references such as Identity matrix ($\mathbf{I}$) and Euclidean mean ($\overline{\mathbf{C}}_E$) \cite{barachant2011multiclass,barachant2013classification}. The Euclidean distance, Euclidean mean, and Riemannian mean equations are presented as following:
\begin{equation}\label{Euclidean_distance}
\delta_E(\mathbf{C}, \mathbf{C}_i) = ||\mathbf{C}- \mathbf{C}_i||_F, 
\end{equation} where $||.||_F$ is the Frobenius norm of a matrix.
\begin{equation}\label{Euclidean_mean}
\overline{\mathbf{C}}_E = \underset{\mathbf{C}}{\arg\min}{(\sum_{i=1}^P \delta_E^2 ( \mathbf{C} , \mathbf{C}_i))} = \frac{1}{P}\sum_{i=1}^P\mathbf{C}_i,
\end{equation}
\begin{equation}\label{riemannian_mean}
\overline{\mathbf{C}}_R = \underset{\mathbf{C}}{\arg\min}{(\sum_{i=1}^P \delta_R^2 ( \mathbf{C} , \mathbf{C}_i))}.
\end{equation}

Accordingly, we employ the Riemannian mean as $\mathbf{C}_{\cref}$. Since there are no closed-form solutions for computing Riemannian mean, we implement the gradient descent algorithm (Algorithm \ref{Riemannian Mean Algorithm}) presented in \cite{fletcher2004principal} for its efficient computation \cite{barachant2011multiclass}. The algorithm implements an iterative procedure to approximate Riemannian mean through minimizing the arithmetic mean of the tangent vectors $\mathbf{J}$. In the first step, we initialize $\mathbf{C}_{\cref}$ using arithmetic mean. Then we use Logarithm mapping to project $\mathbf{C}_i$ to the tangent space $\mathbf{T}_\mathbf{C} \mathcal{M}$ and compute $\mathbf{J}$. Next, we project $\mathbf{J}$ back to $\mathcal{M}$ to update $\mathbf{C}_{\cref}$. The algorithm terminates if either Frobenius norm of $\mathbf{J}$ is less than the tolerance value ($\epsilon=10^{-9}$) or the algorithm reaches maximum iteration of $50$ times. Lastly, we use Riemannian mean as $\mathbf{C}_{\cref}$ to project geodesic information of $\{\mathbf{C}_i\}_{i=1}^P$ onto the same tangent space \cite{tuzel2008pedestrian}. 

\begin{algorithm}[!t]
\caption{Riemannian Mean Algorithm} \label{Riemannian Mean Algorithm}
\begin{algorithmic}[1]
\Procedure {Estimation}{$\mathbf{C}_{\cref}$}
\State $C_{\cref}$ Initialization: Eq. \ref{Euclidean_mean}
\Repeat
\State $\mathbf{J} = \frac{1}{P}\sum_{i=1}^{P}\Log_{C_{\cref}}(\mathbf{C}_i)$
\State $\mathbf{C}_{\cref} = \Exp_{\mathbf{C}_{\cref}}(\mathbf{J})$
\Until{$||\mathbf{J}||_F< \epsilon$}
\State \Return {$\mathbf{C}_{\cref}$}

\EndProcedure
\end{algorithmic}
\end{algorithm}

\subsubsection{Tangent Space Learning for $\{\mathbf{C}_i\}_{i=1}^P$ in $S_R^{+}$}
As mentioned in the previous section, we obtain the geodesic information ($||\mathbf{T}_i||_{\mathbf{C}_{\cref}}$) between $\mathbf{C}_i$ and $\mathbf{C}_{\cref}$ in $S_R^{+}$ based on Logarithm mapping (Eq. \ref{logarithm map}) using the estimated Riemannian mean (Algorithm \ref{Riemannian Mean Algorithm}) as $\mathbf{C}_{\cref}$. Then, in order to obtain the feature vectors containing geodesic information in $\mathcal{M}$ for classification or regression purposes, we require a mapping $\phi_{\mathbf{C}_{\cref}}$: $\mathbf{T}_{\mathbf{C}_{\cref}}\mathcal{M}\rightarrow \mathbb{R}^{R\times(R+1)/2}$, such that $\forall \mathbf{T}_i\in \mathbf{T}_{\mathbf{C_{\cref}}}{\mathcal{M}}$, $||\mathbf{T}_i||_{\mathbf{C}_{\cref}}= ||\phi_{\mathbf{C}_{\cref}}(\mathbf{T}_i)||_2$ \cite{sabbagh2019manifold}. From Eq. \ref{inner_product_norm} and \ref{logarithm map}, we have:
\begin{equation}\label{norm}\small
\begin{split}
||\mathbf{T}_i||_{\mathbf{C}_{\cref}} = \Big[\Tr \big[\Log_{\mathbf{C}_{\cref}}(\mathbf{C}_i)\mathbf{C}_{\cref}^{-1} \Log_{\mathbf{C}_{\cref}}(\mathbf{C}_i)\mathbf{C}_{\cref}^{-1}\big]\Big]^{1/2}\\
     = \Big[\Tr \big[\log(\mathbf{C}_{\cref}^{-1/2}\mathbf{C}_i\mathbf{C}_{\cref}^{-1/2})\log(\mathbf{C}_{\cref}^{-1/2}\mathbf{C}_i\mathbf{C}_{\cref}^{-1/2}) \big]\Big]^{1/2}\\
     = ||\mathbf{S}_i||_F = ||\Vect(\mathbf{S}_i)||_2,   
\end{split}
\end{equation}
where $\Vect{(.)}$ is the vectorization operator and  $\mathbf{S}_i=\log(\mathbf{C}_{\cref}^{-1/2}\mathbf{C}_i\mathbf{C}_{\cref}^{-1/2})$. Therefore, from Eq. \ref{norm}, we have $\phi_{\mathbf{C}_{\cref}}(\mathbf{T}_i)= \Vect(\mathbf{S}_i)$.

\subsubsection{Vectorization}
To further present the geodesic information of $\mathbf{C}_i$ in $\mathcal{M}$ as spatial feature vectors, we denote the spatial information mapping as $\Phi_{\mathbf{C}_{\cref}}$: $\mathcal{M} \rightarrow \mathbb{R}^{R\times (R+1)/2}$, such that $\forall \mathbf{C}_i\in \mathcal{M}$, $\Phi_{\mathbf{C}_{\cref}}(\mathbf{C}_i) =\phi_{\mathbf{C_{\cref}}}({\Log_\mathbf{C_{\cref}}(\mathbf{C}_i))}$. From Eq. \ref{distance_definition} and \ref{norm}, we obtain $\delta_R(\mathbf{C}_{\cref}, \mathbf{C}_i) = ||\Phi_{\mathbf{C}_{\cref}}(\mathbf{C}_i)||_2 =  ||\Vect(\mathbf{S}_i)||_2$. Furthermore, if $\{\mathbf{C}_i\}_{i=1}^P$ are in a small region on $\mathcal{M}$ as stated in \cite{barachant2011multiclass,sabbagh2019manifold}, we have:
\begin{equation}\label{distance approximation}
\delta_R(\mathbf{C}_i, \mathbf{C}_j) \approx ||\Phi_{\mathbf{C}_{\cref}}(\mathbf{C}_i)-\Phi_{\mathbf{C}_{\cref}}(\mathbf{C}_j)||_2, i\neq j, \forall i,j \in [1,P],
\end{equation} where $\mathbf{C}_{\cref}$ is the Riemannian mean of $\{\mathbf{C}_i\}_{i=1}^P$. Eq. \ref{distance approximation} demonstrates that the geodesic distance between $\mathbf{C}_i$ and $\mathbf{C}_j$ can be approximated by the geodesic distance between $\{\mathbf{C}_i\}_{i=1}^P$ and Riemannian mean. Therefore, the geodesic information obtained through the tangent space learning using Riemannian mean as reference are able to represent geodesic information for $\{\mathbf{C}_i\}_{i=1}^P$.
Important information of $\Phi_{\mathbf{C}_{\cref}}(\mathbf{C}_i)$ are completely determined by upper triangular components of $\mathbf{S}_i$ since it is symmetric. Thus, we use the half-vectorization $\Upper(\mathbf{S}_i)$ instead of full-vectorization $\Vect(\mathbf{S}_i)$ to represent $\Phi_{\mathbf{C}_{\cref}}(\mathbf{C}_i)$. As suggested in \cite{barachant2013classification}, we apply coefficient of $\sqrt{2}$ on off-diagonal elements, in order to maintain equality brought by norms $||\mathbf{S}_i||_F = ||\Upper(\mathbf{S}_i)||_2$ (also as in Eq. \ref{norm}), where $\Upper(\mathbf{S}_i)$ = [$\mathbf{\mathbf{S}_i}_{1,1}, ..., \sqrt{2}\mathbf{\mathbf{S}_i}_{1,R}; \mathbf{\mathbf{S}_i}_{2,2}, ...,\sqrt{2}\mathbf{\mathbf{S}_i}_{2,R};...;\mathbf{\mathbf{S}_i}_{R,R}]\in\mathbb{R}^{R(R+1)/2}$. We implement two FC layers to learn the features vectors $\Upper(\mathbf{\mathbf{S}_i})$ concatenated from different frequency sub-bands, as shown in Figure \ref{new_architecture}. To this end, we obtain the spatial information from $\mathbf{C}_i$ on Riemannian manifold as feature vectors $\Phi_{\mathbf{C}_{\cref}}(\mathbf{C}_i)$ in Euclidean space through establishing an information vectorization mapping to preserve local structures such as geodesic information of $\{\mathbf{C}_i\}_{i=1}^P$ in the Riemannian manifold.

\subsection{Fusion Strategy}\label{sec: fusion strategy}
The strategy for the fusion of spatial and temporal information plays an essential role in dealing with multimodal or multi-learning approaches of one modality in order to perform classification/regression. Attention mechanisms have been successfully implemented for refining fusion weights applied to different modalities. For instance, for EEG and electrooculogram (EOG) representation learning, a CapsNet was used in \cite{zhang2021capsule} as an attention mechanism for fusion. 

In the context of this problem, various tasks rely differently on spatial and temporal information. Therefore, a fusion strategy presented as Figure \ref{new_architecture} is adopted. This strategy is inspired by \cite{gu2018hybrid} where a hybrid attention-based multimodal architecture was proposed to learn acoustic and textual features and achieved the state-of-the-art performance on several spoken language classification tasks \cite{gu2018hybrid}. In our architecture, we first use encoders to learn embedding-specific features. Then we employ soft attention to learn the weight ($\alpha$) applied on each embedding-specific feature. Next, we compute the new weighted embedding by multiplying the weight score with the original individual learning embedding. The weighted score of ($1+\alpha$) is adopted to apply the learned weight in order to maintain the original characteristic \cite{gu2018hybrid}. Finally, we perform decision-level fusion on the concatenation of the two new embeddings using an FC layer equipped with different activation functions with respect to discrepant tasks, as illustrated in Figures \ref{new_architecture}.

\begin{table*}[!ht]
\centering
\setlength\tabcolsep{9.0pt}
\caption{Implementation Details for All Four Datasets.}\label{table:implemenation}
\begin{tabularx}{1.92\columnwidth}{c|c|c|c|c|c|c|c|c}
  
    \hline
     Dataset     & \multicolumn{2}{c|}{Filter Bank} &{EEG Trial}  &\multicolumn{2}{c|}{Temporal Information Stream} & \multicolumn{3}{c}{Spatial Information Stream} \\
     \hline
     Dataset     & $H$ & Range     & $T$ & $L$ & Features No. & $N$  & Best Rank     & Features No.\\ 
     \hline \hline
     SEED         &  $5$     & $1.0-50.0Hz$       & $8s$     & $15$     & $10\times62$     & $62$  & $48$     & $5\times48\times(48+1)/2$     \\
     SEED-VIG     &  $25$    & $0.5-50.5Hz$       & $8s$     & $15$     & $50\times17$     & $17$  & $11$     & $25\times 11\times(11+1)/2$   \\
     BCI-IV 2a    &  $25$    & $0.5-50.5Hz$       & $4s$     & $7$      & $50\times22$     & $22$  & $18$     & $25\times18\times(18+1)/2$    \\
     BCI-IV 2b    &  $25$    & $0.5-50.5Hz$       & $4s$     & $7$      & $50\times3$      & $3$   & $3$      & $25\times3\times(3+1)/2$      \\
	\hline
\end{tabularx}
\end{table*}

\section{Experiments}\label{sec: experiments}
\subsection{Datasets}
In the following sections we describe the four datasets used in this study. The EEG data in these datasets have all been recorded with the international $10-20$ system. 

\subsubsection{SEED}
The SEED dataset has been collected as described in \cite{zheng2015investigating} to perform three emotion classification tasks (positive, neutral, and negative). $15$ film clips were chosen as stimuli in the experiments. $15$ subjects ($8$ females and $7$ males, with an average age of $23.3\pm 2.4$) participated in the experiments. Each subject performed experiments in two runs of experiment with $15$ sessions in each run, yielding a total of $30$ sessions. Each session includes four stages: $5$ seconds notice before the movie starts, around $4$ minutes of movie watching, $45$ seconds of self-assessment, and $15$ seconds of rest. $62$ EEG channels were recorded at a sampling frequency of $1000Hz$. EEG signals are split into EEG segments of $T = 8$ seconds with no overlap, as presented in Table \ref{table:implemenation}.

\subsubsection{SEED-VIG}
The SEED-VIG dataset has been collected by \cite{zheng2017multimodal} to estimate driver vigilance. A total of $23$ subjects ($12$ female and $11$ male, with an average age of $23.3\pm 1.4$) participated in the experiment. $17$ channels EEG were collected at sampling frequency of $1000Hz$. The duration of each experiment was around $2$ hours, yielding $885$ EEG trials in total. subjects were asked to drive the simulated car in a virtual environment. Most experiments were performed after lunch so that fatigue during simulated driving could be easily induced \cite{zheng2017multimodal}. The vigilance estimation annotation used a metric called PERCLOS \cite{zheng2017multimodal}, which was measured using eye-tracking glasses. Similar to the SEED dataset, EEG signals are split into EEG segments of $8$ seconds, with no overlap, as shown in Table \ref{table:implemenation}.

\subsubsection{BCI-IV 2a}
The BCI-IV 2a dataset has been collected by \cite{brunner2008bci} to classify four MI tasks (left hand, right hand, tongue and both feet). Each of $9$ subjects ($4$ female and $5$ male, with an average age of $23.1\pm 2.6$) participated experiments in two session on two different days. Each session contain 72 trials for each of the four classes, yielding $288$ trials in total. All sessions contain data without feedback. In these sessions, each trial length is $7.5$ seconds including fixation, visual cue and MI period, and rest. $22$ EEG channels were recorded at the sampling frequency of $250 Hz$ during the experiment. We use the interval of $[2.0-6.0s]$ in each trial ($T=4s$), as presented in Table \ref{table:implemenation}.

\subsubsection{BCI-IV 2b}
The BCI-IV 2b dataset has been collected by \cite{leeb2008bci} to perform binary MI classification (left hand versus right hand). Each of the $9$ subjects ($4$ female and $5$ male, with an average age of $24.2\pm 3.7$) participated in five sessions of the experiment. The first two sessions were conducted without feedback while rest three sessions were conducted with feedback. Each of the first two sessions contain $120$ trials and each of last three sessions contain $160$ trials. In the sessions containing data without feedback, each trial length is $8$ seconds including fixation, visual cue, MI period, and rest. In the sessions containing data with feedback, each trial length is $8$ seconds including visual cue, feedback period, and rest. $3$ EEG channels were recorded at a sampling frequency of $250 Hz$. We use the interval of $[3.4-7.4s]$ in each trial ($T=4s$), as presented in Table \ref{table:implemenation}.

\subsection{Implementation details}

\subsubsection{Filter Bank} 
A filter bank, consisting of $5^{th}$ order Butterworth bandpass filters, was used to decompose EEG signals into different frequency sub-bands. Following that, EEG features were extracted from each of the frequency sub-bands to capture information containing different functional characteristics. For the SEED dataset, DE and logarithm PSD features were calculated on the STFT outputs from five important EEG rhythms, notably delta ($1-3 Hz$), theta ($4-7 Hz$), alpha ($8-13 Hz$), beta ($14-30 Hz$), and gamma ($31-50 Hz$) bands \cite{zheng2015investigating}. For SEED-VIG, DE and logarithm PSD features were determined on the STFT outputs in the range of ($0.5-50.5$ \textit{Hz}) with a $2$ \textit{Hz} resolution \cite{zheng2017multimodal}, yielding a total of $50$ frequency sub-bands. For BCI-IV 2a and BCI-IV 2b datasets, no such standard feature extraction approaches were suggested. We therefore use H=$25$ since our experiments showed that higher resolution bands were beneficial, as shown in Table \ref{table:implemenation}.

\subsubsection{Temporal Information Stream} 
The total number of DE and logarithm PSD features extracted from each of the Hanning windows is $2\times H\times N$, as presented in Table \ref{table:implemenation}. These extracted features are then fed to our attention based LSTM network. Dropout rates of $0.2$, $0.1$, and $0.1$ are applied after each of three LSTM layers with $256$ hidden units respectively to reduce overfitting in BCI-IV 2a dataset, as shown in Figure \ref{new_architecture}. For the rest of datasets, batch normalization (BatchNorm) is applied after each LSTM layer to accelerate the training phase. BatchNorm layers reduce the covariance shift of LSTM output values in each batch, thus increasing model stability \cite{ioffe2015batch}. Then, LeakyReLu (slope of $0.3$) is adopted to enable the activation of hidden neurons for the BatchNorm layer's negative output values \cite{maas2013rectifier}, as shown in Figure \ref{new_architecture}. An FC layer of $64$ units is used for learning temporal information embedding before fusion.

\begin{table*}[!ht]
\centering
\caption{Comparison of Different Solutions and Results for the SEED Dataset.}\label{table:SEED}
\setlength\tabcolsep{18.0pt}
\begin{tabularx}{1.55\columnwidth}{lllll}
    \hline
     Paper & Year  & Input & Method & Acc.$\pm$SD \\
	\hline
	\hline
    Zheng and Lu \cite{zheng2015investigating} & 2015        & DE        & SVM                      & $0.8399\pm 0.0972$ \\
    Zheng and Lu \cite{zheng2015investigating} & 2015        & DE        & DBN                      & $0.8608\pm 0.0834$ \\
    Zheng \cite{zheng2016multichannel}         & 2017        & DE        & GSCCA                    & $0.8296\pm 0.0995$ \\
    Zheng et al. \cite{zheng2017identifying}    & 2017        & DE        & GELM                     & $0.9107\pm 0.0754$ \\
    Zhang et al. \cite{zhang2018spatial}        & 2018        & DE        & STRNN                    & $0.8950\pm 0.0763$ \\
    Song et al. \cite{song2018eeg}              & 2018        & DE        & DGCNN                    & $0.9040\pm 0.0849$ \\
    Li et al. \cite{li2018novel}                & 2018        & DE        & BiDANN                   & $0.9238\pm 0.0704$ \\
    Li et al. \cite{li2019novel}            & 2019        & DE        & BiHDM                    & $0.9312\pm 0.0606$ \\
    Li et al. \cite{li2019eeg}              & 2019        & DE        & GRSLR                    & $0.8841\pm 0.0821$ \\
    Li et al. \cite{li2019regional}         & 2019        & DE        & R2G-STNN                 & $0.9338\pm 0.0596$ \\
    Zhong et al. \cite{zhong2020eeg}           & 2020        & DE        & RGNN                     & $0.9424\pm 0.0595$ \\
    Zhang et al. \cite{zhang2020variational}   & 2020        & DE        & VPR                      & $0.9430\pm 0.0650$ \\

    \textcolor{black}{Li et al. \cite{li2022gmss}} & \textcolor{black}{2022} & \textcolor{black}{DE} & \textcolor{black}{GMSS} & \textcolor{black}{$0.9648\pm 0.0463$} \\
    \hline
	{Ours}                 & \textcolor{black}{2023}        & SCMs, DE, PSD & Ours            & $0.9372\pm 0.0571$ \\
	\hline
\end{tabularx}
\end{table*}

\begin{table*}[!ht]
\centering
\caption{Comparison of Different Solutions and Results for the SEED-VIG Dataset.}\label{table:SEED-VIG}
\setlength\tabcolsep{10.0pt}
\begin{tabularx}{1.6\columnwidth}{lccccc}
    \hline
     Paper & Year & Input  & Method & RMSE$\pm$SD & PCC$\pm$SD\\
	\hline
	\hline
	Huo et al. \cite{huo2016driving}               & 2016     & DE         & GELM                   & $0.1037\pm 0.0309$ & $0.7013\pm 0.1045$ \\	
	Zhang et al. \cite{zhang2016continuous}          & 2016     & DE         & LSTM                   & $0.0927\pm 0.0259$ & $0.8237\pm 0.0831$ \\
	Zheng and Lu \cite{zheng2017multimodal}          & 2017     & DE         & SVR                    & $0.1327\pm 0.0303$ & $0.7001\pm 0.2250$ \\
	Wu et al. \cite{wu2018regression}             & 2018     & DE         & DNNSN                  & $0.1175\pm 0.0420$ & $0.7201\pm 0.1706$ \\
	
        \textcolor{black}{Cheng et al. \cite{cheng2022vigilancenet}} & \textcolor{black}{2022} & \textcolor{black}{DE} & \textcolor{black}{Transformer} & \textcolor{black}{$0.0870\pm 0.0290$} & \textcolor{black}{$0.8970\pm 0.0660$} \\
 \hline
	{Ours}                       & \textcolor{black}{2023}     & SCMs, DE, PSD & Ours           &  ${0.0348\pm 0.0265}$ & ${0.9890\pm 0.0081}$ \\
	\hline
\end{tabularx}
\end{table*}

\subsubsection{Spatial Information Stream} 
Total number of spatial information features in concatenated feature vectors $\Phi_{\mathbf{C}_{\cref}}(\mathbf{C}_i)$ from all the frequency sub-bands is $H\times R\times(R+1)/2$, as presented in Table \ref{table:implemenation}. 
Following the spatial information embedding, a dropout rate of $0.5$ was applied on each of two FC layers consisting of $512$ and $64$ hidden units to prevent overfitting, as shown in Figure \ref{new_architecture}.

\subsubsection{Fusion Strategy} 
As shown in Figure \ref{new_architecture}, each of the two encoders used in the feature fusion block contains an FC layer of $32$ units followed by a single-unit FC layer. The two encoders learn the temporal- and spatial-specific features respectively, as mentioned earlier in Section \ref{sec: fusion strategy}. Lastly, successive to the employed soft-attention mechanism, an FC layer with $128$ units is used to learn the fused and weighted embeddings for decision making.

\subsubsection{Loss Function and Training} 
The loss function of the model and the activation function of the output layer (the final FC in the model as shown in Figure \ref{new_architecture} have been chosen with respect to different task. Since the different datasets involve different classification or regression tasks, different activation function were selected accordingly. Particularly, softmax, sigmoid, softmax and sigmoid were used for SEED, SEED-VIG, BCI-IV 2a, and BCI-IV 2b dataset respectively. Moreover, loss functions were selected with consideration of the different tasks and activation functions. Specifically, categorical cross-entropy, mean squared error, categorical cross-entropy and binary cross-entropy were used for the 4 datasets respectively. Adam optimizer \cite{kingma2014adam} with default learning rate is used to help minimize the loss. \textcolor{black}{We use $200$ epochs and batch size of $32$ to train our network.} The pipeline is implemented using TensorFlow on a pair of NVIDIA RTX $2080$Ti GPUs.

\textcolor{black}{For hyper-parameter selection, we use the training set of the cross-validation scheme (which we use for training and validation). Specifically, we used $20\%$ of training data as validation set for tuning hyperparameter $R$. Regarding other hyperparameters such as learning rate, batch size and number of LSTM layers, we chose to rely on standards used in existing methods of related work, instead of further tuning.}. Moreover, for the spatial information stream, the Riemannian mean of the test set is chosen only with the trials in each batch rather than with the entire set.

We make our source code publicly available at \href{https://github.com/guangyizhangbci/EEG\_Riemannian}{https://github.com/guangyizhangbci/EEG\_Riemannian}.

\begin{table*}[t!]
\centering
\setlength\tabcolsep{13pt}
\caption{Comparison of Different Solutions and Results for the BCI-IV 2a Dataset.}\label{table:BCI-IV 2a}
\begin{tabularx}{1.7\columnwidth}{llllll}
    \hline
     Paper & Year & Input & Method & K/Acc.$\pm$SD \\
	\hline
	\hline
	Tangermann et al. \cite{tangermann2012review}           & 2012    & SCMs           & CSP + NB                  & K:  $\enspace$ $0.5700\pm 0.1830$ \\
    Ang et al. \cite{ang2012filter}                         & 2012    & SCMs           & FBCSP + NB                & K:  $\enspace$ $0.5720\pm 0.2123$ \\
    Ghaheri and Ahmadyfard \cite{ghaheri2013extracting}     & 2013    & SCMs           & CSP + LDA                 & K:  $\enspace$ $0.6156\pm 0.1961$ \\
    Sakhavi et al. \cite{sakhavi2015parallel}               & 2015    & SCMs           & CSP + CNN, MLP            & Acc.:$0.7060\pm 0.1560$  \\	
    Gaur et al. \cite{gaur2018multi}                        & 2018    & SCMs           & MEMD                      & K:  $\enspace$ $0.6011\pm 0.2273$ \\
    Sakhavi et al. \cite{sakhavi2018learning}               & 2018    & SCMs, Raw EEG   & CNN                       & K:  $\enspace$ $0.6594\pm 0.2044$ \\
    Sakhavi et al. \cite{sakhavi2018learning}               & 2018    & SCMs, Raw EEG   & CNN                       & Acc.:$0.7446\pm 0.1533$  \\
    Li et al. \cite{li2019motor}                            & 2019    & SCMs           & LIE + SVM                 & K:  $\enspace$ $0.5633\pm 0.2128$ \\
    \textcolor{black}{Fumanal-Idocin et al. \cite{fumanal2021motor}} & \textcolor{black}{2021}    & \textcolor{black}{SCMs}  & \textcolor{black}{CSP + Fusion}  & \textcolor{black}{Acc.:$0.8540\pm 0.0303$} \\
    \textcolor{black}{Altaheri et al. \cite{altaheri2022physics}} & \textcolor{black}{2022}    & \textcolor{black}{Raw EEG}  & \textcolor{black}{Att. + CNN}    & \textcolor{black}{ K:  $\enspace$ $0.8100\pm 0.1200$} \\
    \textcolor{black}{Altaheri et al. \cite{altaheri2022physics}} & \textcolor{black}{2022}    & \textcolor{black}{Raw EEG}  & \textcolor{black}{Att. + CNN}    & \textcolor{black}{Acc.:$0.8540\pm 0.0910$} \\

    \hline
	\multirow{2}{*}{Ours} & \multirow{2}{*}{\textcolor{black}{2023}}  & \multirow{2}{*}{SCMs, DE, PSD} & \multirow{2}{*}{Ours}    &  K: $\enspace$ ${0.6734\pm0.1381}$ \\
	&   &  &    & Acc.:${0.7551\pm 0.1058}$ \\
	\hline
\end{tabularx}
\end{table*}

\begin{table*}[t!]
\centering
\setlength\tabcolsep{15.0pt}
\caption{Comparison of Different Solutions and Results for the BCI-IV 2b Dataset.}\label{table:BCI-IV 2b}
\begin{tabularx}{1.65\columnwidth}{lllll}
    \hline
     Paper & Year & Input  & Method   & K/Acc. $\pm$SD \\
	\hline
	\hline
	Tangermann et al. \cite{tangermann2012review}    & 2012   & SCMs               & FBCSP + NB                                    & K:  $\enspace$ $0.6000\pm 0.2762$ \\
	Bentlemsan et al. \cite{bentlemsan2014random}    & 2014   & SCMs               & FBCSP + RF                                    & K:  $\enspace$ $0.5988\pm 0.2611$ \\
	\multirow{2}{*}{Rong et al. \cite{rong2018incremental}} & \multirow{2}{*}{2018} & \multirow{2}{*}{SCMs} & \multirow{2}{*}{CSP + ESAFIS}  & K:  $\enspace$ $0.6174\pm 0.1822$ \\	
	&  &  &  & Acc.:$0.8090\pm 0.0907$ \\	
	Ha and Jeong \cite{ha2019decoding}              & 2019   & Spectrogram       & CNN                               & Acc.:$0.7499\pm 0.1452$ \\
	Ha and Jeong \cite{ha2019decoding}              & 2019   & Spectrogram       & CapsNet                          & Acc.:$0.7700\pm 0.1472$ \\
        \textcolor{black}{Kim et al. \cite{kim2023bridging}} & \textcolor{black}{2023} & \textcolor{black}{Spectrogram}    & \textcolor{black}{CNN}   &\textcolor{black}{Acc.: $0.8699\pm N/A$}\\
	\hline
	\multirow{2}{*}{{Ours}} & \multirow{2}{*}{\textcolor{black}{2023}} & \multirow{2}{*}{SCMs, DE, PSD}  & \multirow{2}{*}{Ours}  & K:  $\enspace$ ${0.6720\pm 0.2800}$ \\
	                               &        &                      &                 & Acc.:${0.8360\pm 0.1390}$  \\
	\hline
\end{tabularx}
\end{table*}

\subsection{Evaluation Protocol}
To evaluate our architecture, we adopt the same subject-dependent protocols that have been used in the original papers accompanying the datasets. In the following sections we describe the evaluation protocol details and metrics in detail for each dataset.

\subsubsection{SEED} 
As in \cite{zheng2015investigating}, we use the pre-defined $9$ sessions as training data and the remaining $6$ sessions as testing data in each experiment run, yielding $248$ and $170$ EEG trials for training and testing, respectively. 
\subsubsection{SEED-VIG} 
For this dataset, we use $5$-fold cross-validation to split the data into training and testing sets, as in \cite{zheng2017multimodal}. Two frequently used evaluation metrics for regression, notably root mean squared error (RMSE) and Pearson correlation coefficient (PCC) have been used \cite{zheng2017multimodal}.

\subsubsection{BCI-IV 2a} 
As in \cite{brunner2008bci}, we use the pre-defined training and testing data to evaluate our model, where each contains $288$ EEG trials. Both accuracy (Acc.) and kappa values ($K = \frac{P_0-P_e}{1-P_e}$) are used where $P_0$ is the observed agreement ratio (identical to accuracy), and $P_e$ is the expected agreement ratio while labels are assigned randomly. This metric aims to evaluate the agreement between two label vectors.

\subsubsection{BCI-IV 2b} 
As per \cite{leeb2008bci}, we use the pre-defined training data (first three sessions) with a total of $400$ trials and testing data (last two sessions) with a total of $320$ trials to evaluate our model. We also use the same evaluation metrics as in the BCI-IV 2a dataset.

\subsection{Results and Comparison}

\subsubsection{SEED}
Table \ref{table:SEED} shows the performance comparison between our model and other related works on the SEED dataset. We compare our results to the existing methods that include statistical models, machine learning method, and deep learning algorithms. \textcolor{black}{Generally, deep learning techniques outperform classical machine learning methods} (e.g., SVM, KNN, LR \cite{zheng2015investigating}). In \cite{song2018eeg,zhong2020eeg}, DGCNN and RGNN learned the spatial information through discovering the topological structure of EEG channels using graphs, achieving accuracies of $90.40\%$ and $94.24\%$, respectively. In \cite{zhang2018spatial,li2019regional}, STRNN and R2G-STNN utilized both spatial and temporal information with RNN or LSTM to provide performances of $89.50\%$ and $93.38\%$, respectively. In \cite{li2018novel,li2019novel}, BiDANN and BiHDM employed DAN to utilize the prior distribution information of the target domain, achieving very high performances of $92.38\%$ and $93.12\%$ respectively. \textcolor{black}{Recently, in \cite{li2022gmss}, a graph-based multi-task self-supervised (GMSS) learning method learned more general EEG graph representation by integrating self-supervised tasks and contrastive learning tasks, achieving best accuracy of $96.48\%$.
Our model fully explores the spatial and temporal information, approaching the state-of-the-art result.}

\subsubsection{SEED-VIG}
The comparison of our model and other existing work on the SEED-VIG dataset is shown in Table \ref{table:SEED-VIG}. In \cite{zheng2017multimodal} a baseline SVR model obtained an RMSE of $0.1327$ and a PCC of $0.7001$. In \cite{wu2018regression}, DNNSN used subnetwork nodes to process the DE features, achieving an RMSE of $0.1175$ and a PCC of $0.7201$. In \cite{huo2016driving}, GELM outperformed SVR with an RMSE of $0.1037$ and a PCC of $0.7013$. In \cite{zhang2016continuous}, temporal dependency information learned by LSTM provided a considerable results with an RMSE of $0.0927$ and a PCC of $0.8237$. 
\textcolor{black}{In \cite{cheng2022vigilancenet}, a transformer encoder was used to learn the temporal information, obtaining an RMSE of $0.0870$ and a PCC of $0.8970$.}
Our model achieves considerably superior results with an RMSE of $0.0348$ and a PCC of $0.9890$, \textit{setting a new state-of-the-art} for this dataset.

\subsubsection{BCI-IV 2a}
We compare the performance of our architecture on this dataset to other methods as presented in Table \ref{table:BCI-IV 2a}. In \cite{tangermann2012review,ang2012filter,ghaheri2013extracting}, pipelines consisting of CSP or FBCSP as feature extractors, followed by machine learning technique (e.g., NB, LDA) have been implemented. In \cite{gaur2018multi}, a filter method based on multivariate empirical mode decomposition (MEMD) was employed. In \cite{ghaheri2013extracting}, CSP followed by LDA achieves a kappa value of $0.6156$. In \cite{sakhavi2018learning}, the use of a CNN applied to SCMs outperformed the aforementioned pipelines with a kappa of $0.6594$ and accuracy of $0.7446$. In \cite{li2019motor}, pipelines used the LIE approach to extract spatial features, followed by an SVM classifier. Pipelines using CSP as feature extractor and CNN and MLP as classifier achieved an accuracy of $70.60\%$.
\textcolor{black}{In \cite{fumanal2021motor}, a method using CSP for feature extraction followed by an ensemble of classifiers achieved an accuracy of $85.40\%$. In \cite{altaheri2022physics}, an attention-based CNN framework has been applied on raw EEG signals, obtaining a kappa of $0.8100$ and accuracy of $85.40\%$, achieving state-of-the-art results.
Our model achieves a kappa of $0.6734$ and an accuracy of $75.51\%$.} For fair comparison, we do not consider the references that have employed different evaluation protocols on this dataset (e.g., \cite{xie2016motor, hersche2020compressing}). Moreover, we do not compare our results to references that have performed binary classification (e.g. \cite{barachant2013classification}).

\subsubsection{BCI-IV 2b}
Table \ref{table:BCI-IV 2b} presents the results of our method and related works using this dataset. In \cite{tangermann2012review}, FBCSP followed by NB as the classifier shows very good performance with a kappa of $0.6000$, obtaining the first rank in the BCI competition. A very similar method, using an RF instead of the NB achieves very similar results in  \cite{bentlemsan2014random}. In \cite{ha2019decoding}, deep learning techniques such as CNN and CapsNet have been employed to learn the discriminative information from spectrograms instead of SCMs, achieving accuracies of $74.99\%$ and $77.00\%$, respectively. \textcolor{black}{In \cite{rong2018incremental}, a method using CSP for feature extraction and ESAFIS for classification obtained a kappa of $0.6174$ and an accuracy of $80.90\%$. In \cite{kim2023bridging}, a CNN-based transfer learning architecture achieved the best result with an accuracy of $86.99\%$.}

Our framework achieved considerably better results with a kappa of $0.6720$ and an accuracy of $83.60\%$, \textcolor{black}{approaching} \textit{state-of-the-art}. Similar to other BCI-IV 2a, references that use different evaluation protocols (e.g., \cite{luo2016dynamic,sun2018contralateral,li2019densely}) are not listed in this table.

\begin{table*}[!ht]
\centering
\caption{Result Summary for All Four Datasets as Well as the Different Streams within Our Network.}\label{table:Summary}
\begin{tabularx}{1.75\columnwidth}{c|c|c|c|c|c|c|c}
    \hline
     Dataset & SEED  & \multicolumn{2}{c|}{SEED-VIG} & \multicolumn{2}{c|}{BCI-IV 2a}  & \multicolumn{2}{c}{BCI-IV 2b} \\
	\hline
     Metric     & Acc.      & RMSE      & PCC       & K         & Acc.      & K         & Acc.   \\      
     \hline
     \hline
     State-of-the-art       & \textcolor{black}{\cite{li2022gmss}: $0.9648$}  
     & \textcolor{black}{\cite{cheng2022vigilancenet}: $0.0870$} 
     & \textcolor{black}{\cite{cheng2022vigilancenet}: $0.8970$}
     & \textcolor{black}{\cite{altaheri2022physics}: $0.8100$}
     & \textcolor{black}{\cite{altaheri2022physics}: $0.8540$} 
     & \textcolor{black}{\cite{kim2023bridging}: $N/A$}   
     & \textcolor{black}{\cite{kim2023bridging}: $0.8699$} 
     \\
     Temporal   &$0.9240$   &$0.0383$   &$0.9821$   &$0.2291$   &$0.4219$   &$0.6144$   &$0.8073$ \\
     Spatial    &$0.8570$   &$0.0918$   &$0.8830$   &$0.6636$   &$0.7477$   &$0.6217$   &$0.8111$ \\
     Ours      &$0.9372$   &$0.0348$   &$0.9890$   &$0.6734$   &$0.7551$   &$0.6720$   &$0.8360$ \\
	\hline
\end{tabularx}
\end{table*}

\begin{table*}[!ht]
\begin{center}
\centering
\setlength\tabcolsep{2.5pt}
\caption{Impact of Riemannian Approach on Spatial Information Learning.}\label{table:ablation-approach}
\begin{tabularx}{2.04\columnwidth}{c|c|c|c|c|c|c|c}
	\hline
    Dataset         & SEED                  & \multicolumn{2}{c|}{SEED-VIG }          & \multicolumn{2}{c|}{BCI-IV 2a}      & \multicolumn{2}{c}{BCI-IV 2b}      \\
    \hline
    Metric          & Acc.$\pm$SD            & RMSE$\pm$SD       & PCC$\pm$SD        & K$\pm$SD   & Acc.$\pm$SD   & K$\pm$SD    & Acc.$\pm$SD    \\
    \hline\hline
    SCMs+Vect+FC    & $0.7560 \pm 0.1150$   & $0.1593\pm 0.0678$ & $0.6421\pm 0.2193$ & $0.2043\pm 0.1187$ & $0.4031\pm 0.0890$  & $0.6071\pm 0.2767$ & $0.8037\pm 0.1373$  \\ 
    SCMs+CNN         & $0.7771 \pm 0.1231$   & $0.1751\pm 0.0537$ & $0.5423\pm0.2161$ & $0.3820\pm 0.1896$ & $0.5365\pm 0.1430$  & $0.3737\pm 0.1989$ & $0.6869\pm 0.0991$ \\ 
    SCMs+CapsNet     & $0.6853 \pm 0.1407$   & $0.1903\pm 0.0658$ & $0.4763\pm 0.1710$ & $0.2684\pm 0.1558$ & $0.4513\pm 0.1153$  & $0.3412\pm 0.2047$ & $0.6703\pm 0.1021$ \\
    SCMs+Riem.+FC   & $0.8570 \pm 0.0932$   & $0.0918\pm 0.0276$ & $0.8830\pm 0.0849$ & $0.6636\pm 0.1437$ & $0.7477\pm 0.1172$  & $0.6217\pm 0.3001$ & $0.8111\pm 0.1380$ \\ 

	\hline
\end{tabularx}
\end{center}
\end{table*}

\begin{table*}[!ht]
\begin{center}
\centering
\caption{Impact of Different Fusion Methods on Learned Embeddings.}\label{table:fusion}
\setlength\tabcolsep{3pt}
\begin{tabularx}{2.0\columnwidth}{c|c|c|c|c|c|c|c}
	\hline
    Dataset         & SEED                  &\multicolumn{2}{c|}{SEED-VIG}      &\multicolumn{2}{c|}{BCI-IV 2a}          & \multicolumn{2}{c}{BCI-IV 2b}         \\
    \hline
    Metric          & Acc.$\pm$SD           & RMSE$\pm$SD        & PCC$\pm$SD           & K$\pm$SD           & Acc.$\pm$SD         & K$\pm$SD        & Acc.$\pm$SD    \\
    \hline\hline
    Concatenation   & $0.9100\pm 0.0825$    & $0.0355\pm 0.0261$ & $0.9857\pm 0.0091$   & $0.6405\pm 0.1586$ & $0.7304\pm 0.1190$  & $0.6623\pm 0.2520$  & $0.8312\pm 0.1264$ \\    
    Soft attention  & $0.9250\pm 0.0711$    & $0.0350\pm 0.0227$ & $0.9887\pm 0.0089$   & $0.6619\pm 0.1422$ & $0.7464\pm 0.1066$  & $0.6397\pm 0.2600$  & $0.8203\pm 0.1295$ \\ 
    Ours            & $0.9372\pm 0.0571$    & $0.0348\pm 0.0265$ &$0.9890\pm 0.0081$    & $0.6734\pm 0.1381$ & $0.7551\pm 0.1058$  & $0.6720\pm 0.2800$  & $0.8360\pm 0.1390$ \\    
	\hline
\end{tabularx}
\end{center}
\end{table*}

\subsection{Discussion}
Table \ref{table:Summary} presents the summary of the performance of our proposed model compared to the state-of-the-art in the four datasets. We also show the performance of our individual learning streams with the same parameter settings as used in our network. We observe that the spatial information stream performs better in both MI datasets while the temporal information stream performs superior for emotion recognition and vigilance estimation datasets. This demonstrates the necessity to exploit both spatial and temporal information from EEG, in order to develop a generalized model suitable for different BCI applications (e.g., emotion recognition, vigilance estimation, and MI classification). Moreover, we observe that our model achieves much better results than both individual learning streams even when the difference among the performance of the two streams is very small as with the BCI-IV 2b dataset. Interestingly, the performance of our model is only slightly better than each individual stream when the difference between them is large, as seen with the SEED-VIG and BCI-IV 2a datasets. This indicates that the two streams are likely to contain more contradictory information, resulting in difficulty for the model to learn a strong relationship between learned representations and outputs. Overall, the results in Table \ref{table:Summary} show the performance of each stream is dataset-dependent and demonstrate the importance of combining the two streams.

\begin{figure*}
\centering
\begin{minipage}{0.5\textwidth}
   \centering
   \includegraphics[width=1.0\linewidth]{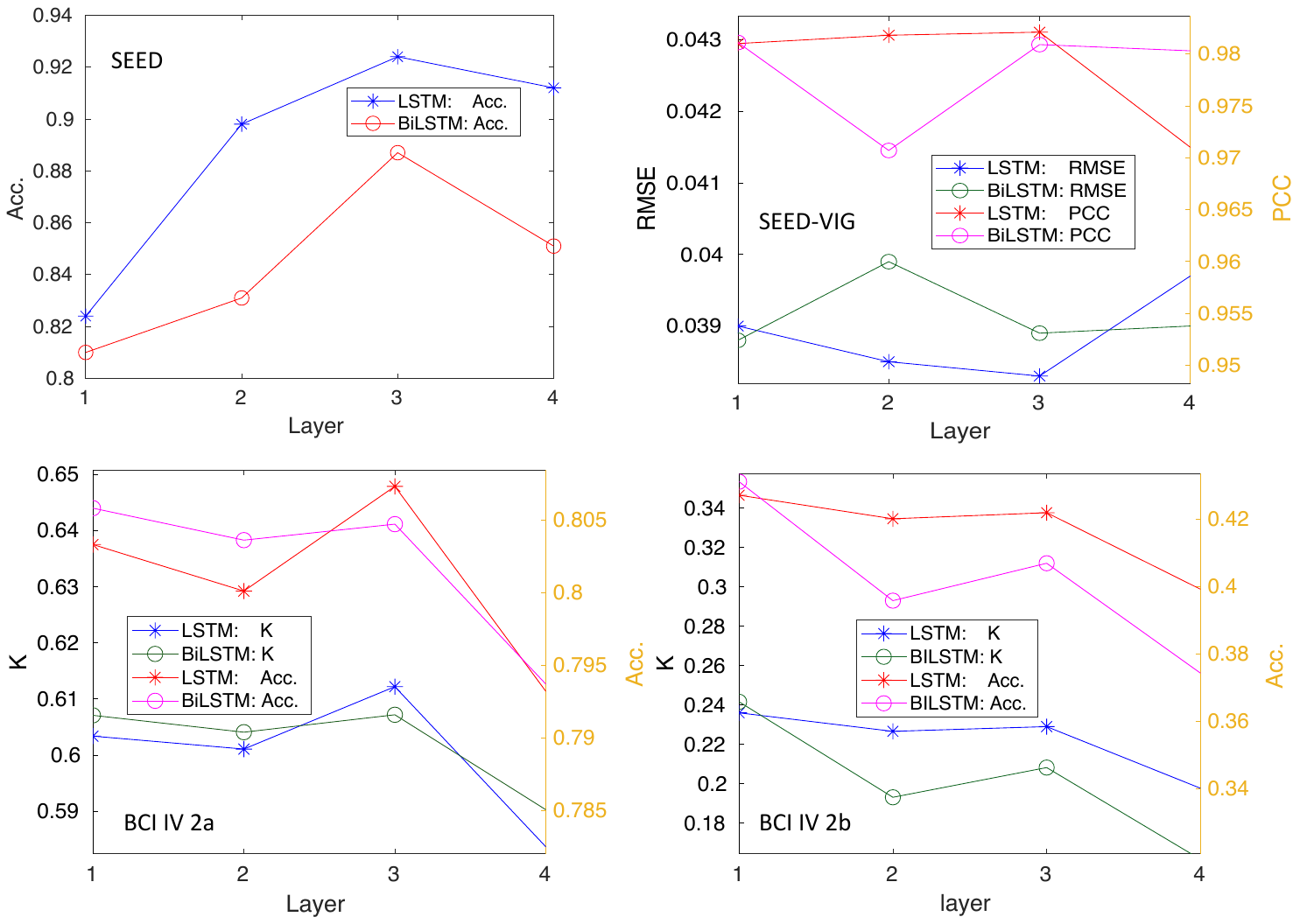}
   \captionof{figure}{Impact of LSTM layers on the temporal stream.}
   \label{fig:ablation_temporal}
\end{minipage}%
\begin{minipage}{0.5\textwidth}
  \centering
  \includegraphics[width=1.0\linewidth]{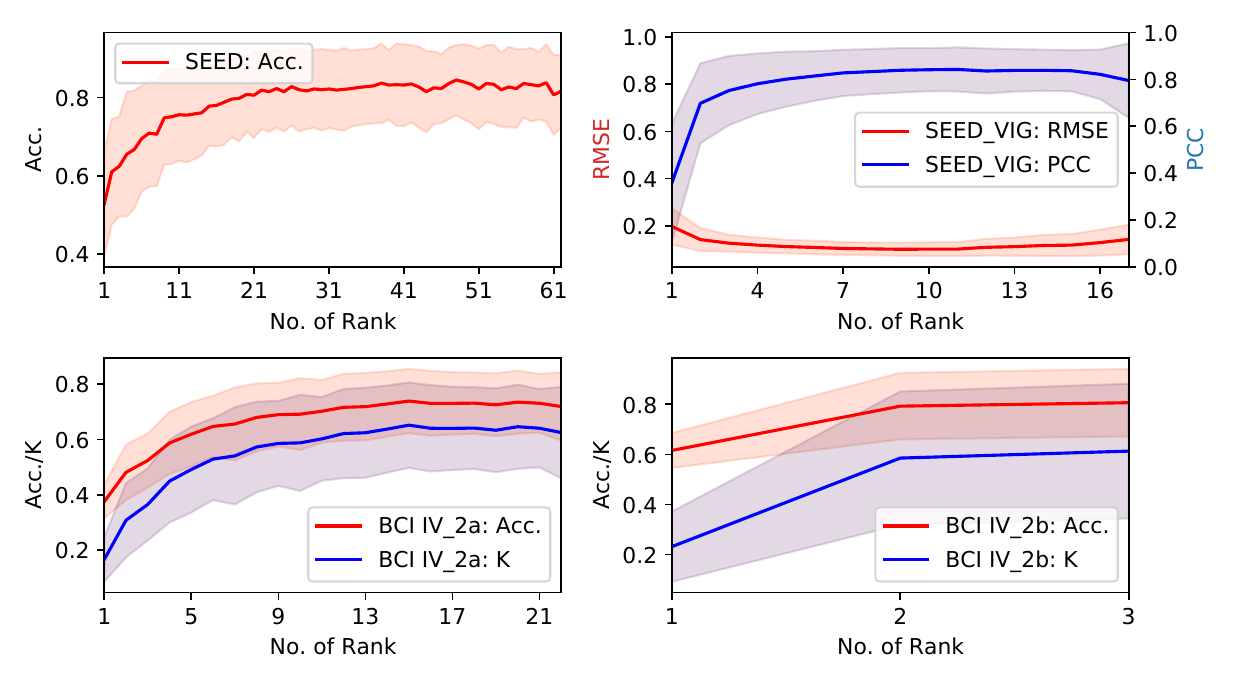}
  \captionof{figure}{Effect of dimensionality reduction on the spatial stream.}
  \label{fig:ablation_spatial}
\end{minipage}
\end{figure*}

\begin{figure*}[ht!]
    \begin{center}
    \includegraphics[width=1.0\textwidth]{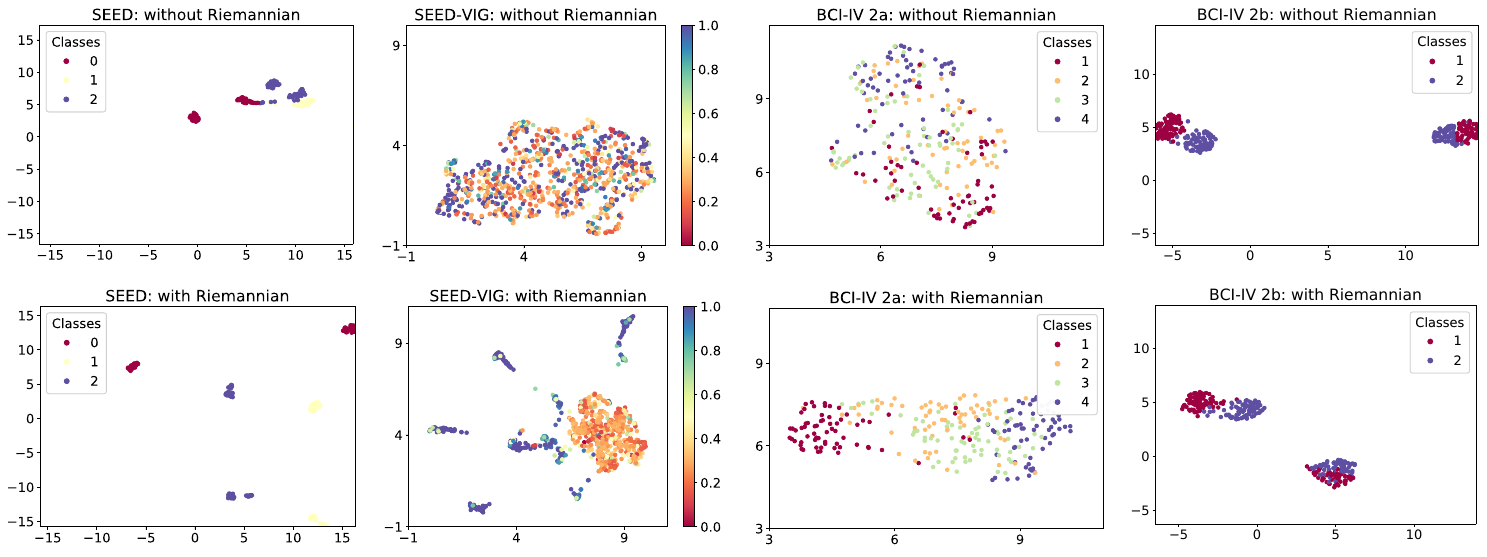} 
    \end{center}
\caption{Comparison between spatial information vectors without Riemannian ($1^{st}$ row) and with Riemannian ($2^{nd}$ row) using UMAP.}
\label{fig:umap}
\end{figure*}

\subsection{Ablation Experiments and Analysis}
We conduct numerous ablation and analysis studies to evaluate the impact of different components of our framework on the performance.

\subsubsection{Impact of LSTM layers on Temporal Information Learning}
We evaluate the depth of the LSTM network and the performance of the LSTM compared with BiLSTM on learning temporal information. As shown in Figure \ref{fig:ablation_temporal}, LSTM with three layers consistently has the best performance among LSTMs with different numbers of layers. Also, the LSTM performs better than Bi-LSTM with the same number of layers for most datasets.

\subsubsection{Importance of Riemannian Approach on Spatial Information Learning}
To show the importance of the Riemannian approach on spatial information learning, we compare our solution with a Euclidean approach that directly employs vectorization followed by FC layers on spatial covariance matrices ($\{\mathbf{C}_i\}_{i=1}^P$). \cite{barachant2013classification}. We also implement other deep learning techniques such as CNN and CapsNet \cite{sabour2017dynamic} directly on $\{\mathbf{C}_i\}_{i=1}^P$ without vectorization for comparison. Table \ref{table:ablation-approach} shows the comparison of these different approaches applied on SCMs for spatial information learning. Our Riemannian approach consistently outperforms other approaches for all 4 datasets, addressing the first challenge of spatial information learning on the Riemannian manifold which was mentioned earlier in the Introduction section.

Next, we explore the learned representation space using uniform manifold approximation and projection (UMAP) \cite{mcinnes2018umap} to better understand the impact of our Riemannian approach. Figure \ref{fig:umap} shows the comparison between the feature spaces using our Riemannian approach $\Phi_{\mathbf{C}_{\cref}}({\{\mathbf{C}_i\}_{i=1}^P})$ versus a direct vectorization of spatial covariance matrices without Riemannian $\Vect({\{\mathbf{C}_i\}_{i=1}^P})$ for a sample subject. In SEED, SEED-VIG, and BCI-IV 2a datasets, the information in $\Phi_{\mathbf{C}_{\cref}}$ with the Riemannian approach are clearly more separable than the information in $\Vect({\{\mathbf{C}_i\}_{i=1}^P})$ without Riemannian. In BCI-IV 2b, the difference in separability is very small. This is likely due to the limited number of channels ($N=3$). Our observations are consistent with the comparison results shown in Table \ref{table:ablation-approach}. Overall, our proposed architecture results in superior separability in the feature space.

\subsubsection{Impact of Dimensionality Reduction on Spatial Information Learning}
We evaluate the effect of dimensionality reduction by observing the performance of spatial information learning with different $R$ values representing the full rank of the covariance matrix. To this end, we perform a grid search on $R$ in the range of $[1, N-1]$. Figure \ref{fig:ablation_spatial} shows the effect of dimensionality reduction with different $R$ values on spatial information learning, based on different evaluation metrics for the 4 datasets. 
We observe that the best performances are achieved at the rank $R$ of $48$, $11$, $18$ for SEED, SEED-VIG, and BCI-IV 2a datasets, respectively. For the BCI-IV 2b dataset, only $3$ EEG channels are available, therefore dimensionality reduction is not necessary, hence, the best performance has been expectedly achieved at $N=3$.
Overall, the results in Figure \ref{fig:ablation_spatial} demonstrate the importance of projecting the SCMs from SPSD to SPD via dimensionality reduction. It also shows that our spatial information approach addressed the second challenge of Riemannian metric learning on SCMs which was mentioned earlier in the Introduction section.

\subsubsection{Impact of Fusion strategy on Both Learning Embeddings}
We employ different feature fusion techniques such as naive concatenation, soft attention mechanisms, and our fusion strategy. The results in Table \ref{table:fusion} show that our method marginally but consistently outperforms other fusion strategies for all the datasets, addressing the third challenge of feature fusion of the Riemannian spatial and Euclidean temporal information which was mentioned in the Introduction section.

\begin{figure}[ht!]
    \begin{center}
    \includegraphics[width=1.0\columnwidth]{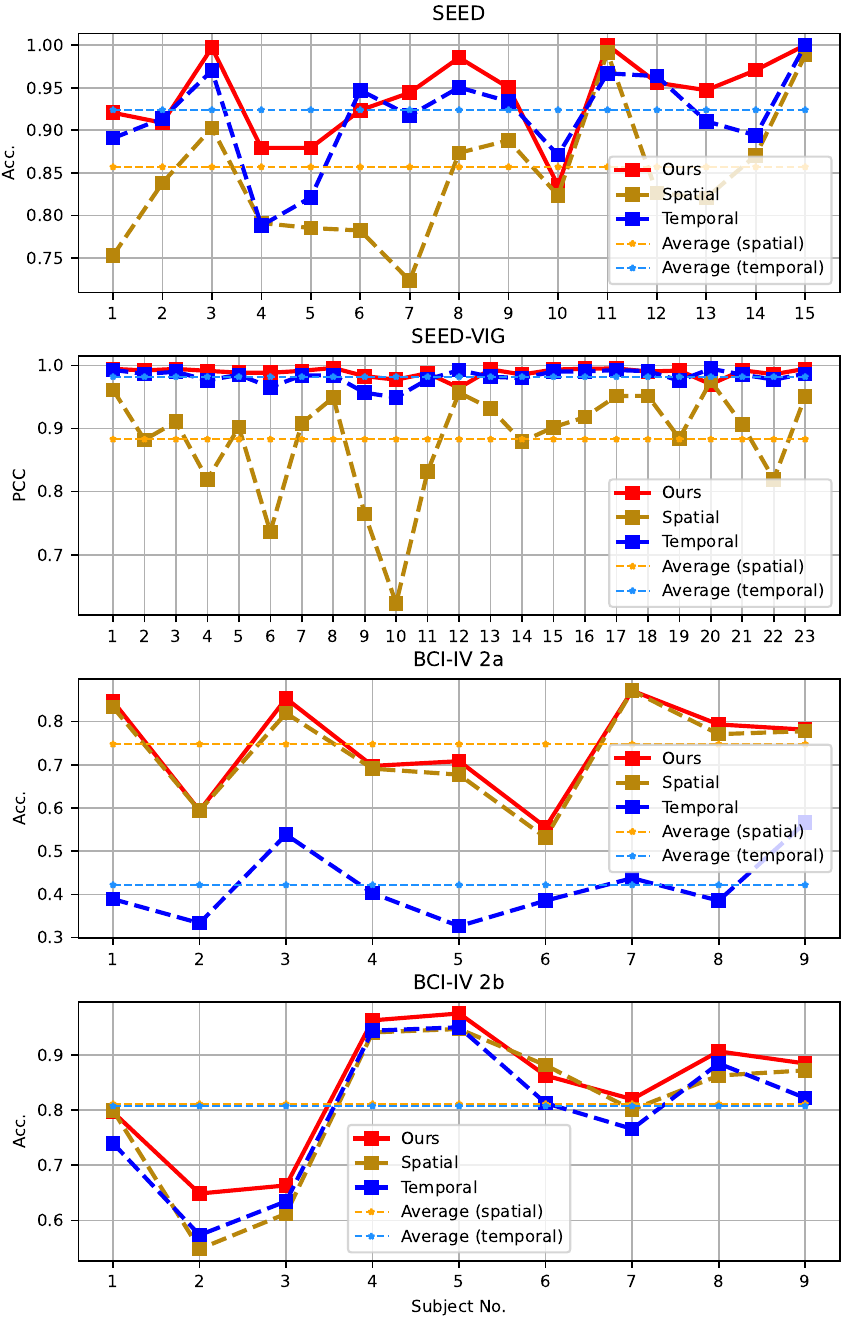} 
    \end{center}
\caption{The performance of our proposed network in comparison to individual spatial and temporal information stream on SEED, SEED-VIG, BCI-IV 2a 
\& 2b datasets.}
\label{fig:subjects}
\end{figure}

\subsubsection{Effect of Our Model on Low-performing Subjects}
We also investigate the impact of our model on subjects with lower-than-average performance (also called low-performing subjects). Figure \ref{fig:subjects} shows the performance of our model compared to individual spatial and temporal information streams on each subject for all the datasets. The figure also presents the average performance of spatial and temporal information streams across all the subjects, which we use to identity the low-performing individuals.
In SEED, for the \textit{spatial information stream}, subjects No. $1,2,4,5,6,7,10,12,13$ are low-performing subjects. It can be observed that our model (fusion of both streams) improves the performance of the lowest performing subject (No. $7$) by $22.06\%$ and the second-lowest performing subject (No. $1$)
by $16.77\%$. The \textit{temporal information stream} on the other hand, shows poor performance on subjects No. $1,2,4,5,7,10,13,14$. Our proposed architecture improves the accuracy by $9.13\%$ on the lowest-performing subject (No. $4$) and $5.82\%$ on the second lowest-performing subject (No. $5$). 
In SEED-VIG, for the spatial information stream, our model has obvious improvements on the low-performing subjects (No. $4,6,9,10,11,22$), particularly by the PCC value of $35.46\%$ on subject No. $10$ (the lowest). For the temporal information stream, the variance of performance across subjects is very small. Our model marginally improves the performance by PCC values of $2.29\%$, $2.62\%$, and $2.81\%$ on the lowest-performing subjects (No. $6$, $9$, and $10$). In BCI-IV 2a, for the spatial information stream, our model shows a small improvement on low-performing subjects (No. $2,4,5,6$), with an improvement of $2.43\%$ on the lowest-performing participant (No. $6$). For the temporal information stream, our model consistently and substantially improves the performance on low-performing subjects (No. $1,2,4,5,6,8$), particularly by the accuracy of $38.19\%$ on the lowest-performance subject (No. $5$). In BCI-IV 2b, we observe that the performance of both information streams on each subjects are comparable. Both streams have the same low-performing subjects (No. $1,2,3,7$). Our network outperforms the spatial and temporal information streams on the lowest-performing subject (No. $2$) by $10.04\%$ and $7.53\%$, respectively. Overall, our model demonstrates its effectiveness by enhancing the performance of subjects with low performance based on individual spatial and temporal information streams.

\section{Conclusions}\label{sec: summary}
In this paper, we propose a novel deep architecture to learn EEG using spatio-temporal information on a Riemannian manifold as well as a Euclidean space. Spatial information is efficiently learned from spatial covariance matrices of EEG signals through our Riemannian approach. Moreover, temporal information is obtained by extracting features from EEG signals in consecutive time periods and learning them using our deep LSTM network followed by an attention mechanism. Our fusion strategy exploits the complementary information from both information streams. We test our framework with four public datasets with various types of tasks in the three popular EEG fields of emotion recognition, vigilance estimation, and MI classification. Our results demonstrate the robustness of our model in both fields on binary classification, multi-class classification, and even regression. \textcolor{black}{We set new state-of-the-art result on SEED-VIG, while approaching the existing state-of-the-art for emotion recognition on the SEED dataset and for MI classification on BCI-IV 2a and BCI-IV 2b datasets}.

\label{sec:refs}


\ifCLASSOPTIONcaptionsoff
  \newpage
\fi



%

\bibliographystyle{IEEEtran}
\bibliography{IEEEabrv,refs}

\begin{IEEEbiographynophoto}{Guangyi (Patrick) Zhang}
is a postdoctoral fellow with the University Health Network, University of Toronto, Canada. His area of research is on AI algorithms for cancer vaccine design. He earned his Ph.D. degree from the Department of Electrical and Computer Engineering, Queen’s University, Canada, where he worked on EEG representation learning for BCI applications. Prior to that, he worked on bio-sensor design and bio-signal processing, at Peking University People’s Hospital in Beijing, China. 
His works have appeared in top venues in his field, such as T-AFFC, T-NSRE, IEEE Sens. J, ICASSP, and ACII. He has also served as a reviewer for T-PAMI, T-AFFC, T-NNLS, T-CYB, T-NSRE, T-ETCI, T-AI, Pattern Recognition, ACII, and ICPR.
\end{IEEEbiographynophoto}

\begin{IEEEbiographynophoto}{Ali Etemad}
is an Associate Professor, as well as a Mitchell Professor in AI for Human Sensing $\&$ Understanding at the Department of Electrical and Computer Engineering, and Ingenuity Labs Research Institute, Queen’s University, Canada. He leads the Ambient Intelligence and Interactive Machines (Aiim) lab, where his main area of research is machine learning and deep learning focused on human-centered applications with wearables, smart devices, and smart environments. 
Dr. Etemad is an Associate Editor for IEEE Transactions on Artificial Intelligence, and has been a PC member/reviewer for many notable conferences and journals in the field. 
He has been the General Chair for the AAAI Workshop on
Representation Learning for 
Responsible Human-Centric AI (2023), General Chair for the AAAI Workshop on Human-Centric Self-Supervised Learning (2022), Publicity Co-Chair for European Workshop on Visual Information Processing (2022), and Industry Relations Chair for Canadian Conference on AI (2019).
Dr. Etemad’s lab and research program have been funded by the Natural Sciences and Engineering Research Council (NSERC) of Canada, Ontario Centers of Excellence (OCE), Canadian Foundation for Innovation (CFI), Mitacs, and other organizations, as well as the private sector.
\end{IEEEbiographynophoto}

\end{document}